# A Survey on Data Augmentation for Text Classification[1]

Markus Bayer[2]

PEASEC², Technical University of Darmstadt, bayer@peasec.tu-darmstadt.de

Marc-André Kaufhold

PEASEC², Technical University of Darmstadt, kaufhold@peasec.tu-darmstadt.de

Christian Reuter

PEASEC², Technical University of Darmstadt, reuter@peasec.tu-darmstadt.de

Data augmentation, the artificial creation of training data for machine learning by transformations, is a widely studied research field across machine learning disciplines. While it is useful for increasing a model's generalization capabilities, it can also address many other challenges and problems, from overcoming a limited amount of training data, to regularizing the objective, to limiting the amount data used to protect privacy. Based on a precise description of the goals and applications of data augmentation and a taxonomy for existing works, this survey is concerned with data augmentation methods for textual classification and aims to provide a concise and comprehensive overview for researchers and practitioners. Derived from the taxonomy, we divide more than 100 methods into 12 different groupings and give state-of-the-art references expounding which methods are highly promising by relating them to each other. Finally, research perspectives that may constitute a building block for future work are provided.

## 1 INTRODUCTION

An increase in training data does not necessarily result in a solution for the learning problem. Nevertheless, the quantity of data remains decisive for the quality of a supervised classifier. Originating from the field of computer vision, many different methods to artificially create such data exist, which are referred to as *data augmentation*. For images, transformations such as rotations or changes of the RGB channel are useful, as the resulting model should be invariant for these. Similar to computer vision, speech recognition uses procedures that change sound or speed. In contrast, research on data augmentation in Natural Language Processing (NLP) faces the difficult task of establishing such universal rules for textual data transformations which, when executed automatically, maintain labeling quality [1], [2].

---

[1] This work has been co-funded by the German Federal Ministry of Education and Research (BMBF) in the project CYWARN (13N15407), by the LOEWE initiative (Hesse, Germany) within the emergenCITY center, as well as the German Federal Ministry of Education and Research and the Hessian Ministry of Higher Education, Research, Science and the Arts within their joint support of the National Research Center for Applied Cybersecurity ATHENE. The calculations for this research were conducted on the Lichtenberg high performance computer of the TU Darmstadt.

[2] Science and Technology for Peace and Security | Pankratiusstraße 2, 64289 Darmstadt, Germany

Research in this area was therefore much more limited before 2019, despite existing extensive areas of application [3]. Nowadays, this challenge remains, but is being addressed by many scientists from different research fields as more possibilities and complex mechanisms open up. Within these fields, researchers strive to meet various goals, e.g., generating more data for low-data regimes, balancing imbalanced dataset classes or securing against adversarial examples. Thus, textual data augmentation comes in many contrasting forms that will be grouped and explained in this survey. We will provide an in-depth analysis and also relate the methods to the state-of-the-art, as they now face another challenge due to the advent of transfer learning. For example, Longpre et al. [4] demonstrate that many data augmentation methods cannot achieve gains when using large pre-trained language models, as they already are invariant to various transformations. They hypothesize that data augmentation methods can only be beneficial, if they create new linguistic patterns that have not been seen before. Keeping this in mind, the survey is closed with a meta-perspective on the methods. This survey is therefore intended to contribute to data augmentation and general text classification by highlighting the following aspects:

- Goals and applications (C1). We highlight the goals and applications of data augmentation that we derive from the comprehensive review. These have only been presented to a limited and incomplete extent in previous research papers.
- Comprehensive survey on data augmentation in text classification (C2). Our survey provides a holistic overview of the data augmentation field in text classification. While methods for other NLP disciplines are mentioned, the listing is not complete, nor are the methods set in relation to each other as the text classification data augmentation methods are.
- Data-structure-driven taxonomy and method-oriented categorization (C3). The text classification data augmentation methods are clustered according to a data-structure-based, high-level taxonomy and then subdivided into more fine-grained method groups. This is also present in the surveys from Shorten and Khoshgoftaar [5] and Wen et al. [6] and is adapted for the text classification domain.
- Method-driven overview and in-depth details (C4). The textual data augmentation methods are explained clearly and concisely while including necessary details for delimitation and comparison. Contrasting to other works, our extensive study contains 12 groups with more than 100 different approaches.
- State-of-the-art review (C5): Within the literature survey we examine latest state-of-the-art considerations, for example, the limited benefit of textual data augmentation methods with large pre-trained models that are often neglected in current works.
- Relating methods (C6). Throughout this survey, the methods are set in relation to conception and performance comparisons, while taking the underlying models and application contexts into account.
- Future research perspectives (C7). We identify future research opportunities that are either necessary for a state-of-the-art comparison or sensible to look into because of current challenges and promising directions for textual data augmentation.

The survey paper is structured as follows: The paper introduces the foundations of data augmentation in Section 2. This section is then broadened by the consideration of the goals and applications. Section 3 is subdivided into the various data augmentation groups and contains the explanations, as well as tabular overviews of the methods. In Section 4, an analysis of the data augmentation methods from a more global perspective is given and various future research directions are discussed. Section 5 outlines the limitations of data augmentation and provides a conclusion for this survey.

## 2 BACKGROUND: FOUNDATIONS, GOALS, AND APPLICATIONS OF DATA AUGMENTATION

In many machine learning scenarios, not enough data is available to train a high-quality classifier. To address this problem, data augmentation can be used. It artificially enlarges the amount of available training data by means of transformations [7]. In the well-known LeNet by LeCun et al. [8], early versions of data augmentation have already been observed. The notion of data augmentation comprises various research in different sub-areas of machine learning. Many scientific works merely relate data augmentation to deep learning, yet it is frequently applied in the entire context of machine learning. Therefore, this paper adopts the notion of data augmentation as a broad concept, encompassing any method that enables the transformation of training data. However, following common understanding in research, semi-supervised learning is not regarded as a form of data augmentation and is only thematized if sensible in this survey.

An important term relating to data augmentation is label preservation, which describes transformations of training data that preserve class information [9]. For example, in sentiment analysis, an entity replacement within a sentence is often sufficient for label preservation, but randomly adding words may alter the sentiment (e.g., an additional "not" could invert the meaning of a sentence). In many research works, label preservation is adapted to also cover transformations changing the class information, if the label is adjusted correctly. Additionally, many transformations do not maintain the correct class in every case, but with a high probability. Shorten and Khoshgoftaar [5] define this probability as the safety of a data augmentation method. When this uncertainty is known, it could be directly integrated in the label. Otherwise, methods like label smoothing [10] can model a general uncertainty.

The goals of data augmentation are manifold and encompass different aspects. As mentioned above, training data is essential for the quality of a supervised machine learning process. Banko and Brill [11] show that only the creation of additional data can improve the quality of a solution in the confusion set disambiguation problem, while the choice of the classifier does not lead to a significant change. The model selection and development will remain a crucial aspect of machine learning. Yet, scholars suggest that in some situations, the choice for higher investments in algorithm-choice and -development instead of corpus-development should be carefully considered [11]. Closely connected to this is the big data wall problem, which Coulombe [9] mentions in his work on data augmentation. It describes that big companies benefit from the special advantage of having access to a large amount of training data. Consequently, the already large GAFAM-Companies (Google/Alphabet, Amazon, Facebook/Meta, Apple, and Microsoft) expand their predominance over smaller businesses due to their data superiority. An ideal data augmentation method could approach these points and decrease the dependency of training data even though full elimination is not likely.

Additionally, creating training data for various classification problems is accompanied by high labeling costs. In many instances, assessment and labeling by experts are necessary to prevent incorrect training examples. These aspects can, for example, be especially stressed concerning the field of crisis informatics [12], [13]. Creating relevant classifiers for emergency services and responders is only possible during crises and requires resources and time from personnel needed elsewhere to, e.g., act as first responders, therefore in the worst-case costing lives [14]. Similarly, training data for medical image processing is very valuable. Due to the rareness of certain diseases, the privacy of patients, and the requirement of medical experts, it is particularly challenging to provide medical datasets [5]. In a related sense, many domains, such as cybersecurity, have a time-critical factor that requires training data to be collected as quickly as possible so that, in terms of the cybersecurity domain example, threats can be responded to quickly [15], [16]. With regard to such classification problems, data augmentation could help minimize the required amount of data needed to be labeled and to solve interlinked problems.

Data augmentation is particularly significant for the field of deep learning. Work such as that by Minaee et al. [17] has already extensively investigated the quality of deep learning algorithms in text classification, but there are many application scenarios where there is not enough data to produce high quality classifiers. For example, Srivastava et al. [18] have also demonstrated that deep neural networks in general are particularly powerful but encompass a tendency to overfit; faced with unseen instances, they might generalize badly. This observation can be illustrated with help of the bias variance dilemma. On the one hand, deep learning algorithms are, due to their deep and non-linear layers, very strong models with a lower bias-error. On the other hand, they show a high variance for different subsets of training data [19]. This problem can be solved by arranging the algorithm to prefer simple solutions or by providing a bigger amount of training data. The first option is aimed at methods of regularization, such as dropout or the addition of a L2 norm via the model's parameters in the loss-function. The second option is frequently realized by means of data augmentation and could, in this context, also be considered as a type of regularization. According to Hernandez-Garcia and König [20], data augmentation is a preferred regularization method, as it achieves generalization without degrading the models' representational capacity and without re-tuning other hyperparameters. While other methods reduce the bias error, data augmentation's objective is to keep it constant and is used to solve the problem at the root [5]. Nonetheless, data augmentation still depends on the underlying classification problem and can therefore not be effectively applied in all circumstances.

In the context of deep learning models, so called adversarial examples/attacks are generated more and more frequently. These small changes in the input data, which are almost unrecognizable to humans, mislead the algorithms to make wrong predictions [21]. Table 1 shows two different genuine examples in which the smallest changes in the texts alter the classification prediction. Alzantot et al. [22] further present an algorithm that generates semantically and syntactically similar instances of training data, successfully outwitting sentiment analysis and entailment models. With the help of adversarial training, these automatic adversarial example generators can be used as data augmentation methods, as done, for example, in [23], [7], [24], or [25], in order for models to be less susceptible to such easy alterations. If the amount of data is taken into consideration, it stands out that certain classification problems are often heavily unbalanced, for instance, only a small amount is relevant (positive) while the irrelevant (negative) data is prevalent [26]. For example, in an entire corpus for topic classification or crisis identification, only few data actually relate to the topics or the crisis in question. Zhong et al. [27] term a dataset as unbalanced, if the distribution of classes within it is not approximately equally balanced. Data augmentation may help to enhance the amount of data for a certain class in order for balanced class distributions to be present and thus for a classifier to be able to be modelled more robustly [28], [29].

Data augmentation can also be helpful in sensitive domains. Dealing with confidential or privacy-related data, one can decrease the usage of real-world data by crafting artificial data. It is even possible to only train the algorithm on the newly created data, in order to prevent drawing any conclusions on non-artificial training data from a deployed model. For example, Carlini et al. [30] have demonstrated a method for extracting training data from large language models that could contain private information. For training such a privacy ensuring model, special data augmentation techniques that are able to anonymize the data have to be used.

Table 1. Examples for Adversarial Attacks adapted from Ebrahimi et al. [21].

| **Original text** | **Altered text** |
|---|---|
| South Africa's historic Soweto township marks its 100th birthday on Tuesday in a mood of optimism.<br>57% **World** | South Africa's historic Soweto township marks its 100th birthday on Tuesday in a moo**P** of optimism.<br>95% **Sci/Tech** |
| Chancellor Gordon Brown has sought to quell speculation over who should run the Labour Party and turned the attack on the opposition Conservatives.<br>75% **World** | Chancellor Gordon Brown has sought to quell speculation over who should run the Labour Party and turned the attack on the o**B**position Conservatives.<br>94% **Business** |

Data augmentation exists in different types and areas of application. A taxonomy of the types in the textual domain can be seen in Figure 1. The augmentation methods can be divided into the transformation of raw data (data space) and processed representations of data (feature space) [5]. These representations are transformed types of data, for example, activation vectors of a neuronal network, the encoding of an Encoder Decoder Network, or LSTM hidden states, respectively embeddings of data. Abstracting from the textual realm, in many cases, data augmentation depends on the underlying problem (text classification, image recognition, etc.); and is therefore applied in different ways in different areas. Procedures generic enough to be used across different areas are for the most part limited to the feature space.

The most substantial research on data augmentation exists in the field of computer vision. This is due to the intuitive construction of simple label preserving transformations. Data augmentation methods in computer vision are, among other, geometric transformations [7], [24], neural style transfers [31]–[33], interpolation of images [34], random partial deletions [35], and generative adversarial network (GAN) data generation [21]. Sophisticated techniques can additively improve the accuracy baseline for different problems by around 10 to 15 percent [35]. Another area of application for data augmentation is speech processing. Researchers have successfully used acoustic transformations of the input data. Ko et al. [36] have achieved up to 4.3 points better accuracy values by modifying speed. Furthermore, interfering with vocal tract length [35] or adding noise [34] may also enhance the quality of classifiers. The application of data augmentation in the textual realm is considered a difficult task, since textual transformations preserving the label are difficult to define [1], [2]. Nevertheless, several simple and sophisticated methods have been developed in this and adjacent research areas.

## 3 TEXTUAL DATA AUGMENTATION METHODS

In the following, different data augmentation methods for textual data are summarized, explained, and subdivided in different groupings. Mainly methods focusing on the application of text classification are included, although augmentation methods for other tasks in the textual realm are also mentioned if they fit into the group. In this survey, text classification is considered a problem of the field of NLP, where units such as sentences, paragraphs, or documents are categorized into class labels [17]. For example, generative or sequence-tagging tasks, where either text has to be generated or the words of the units have to be tagged individually, are not regarded as tasks in this sense. This means that augmentation methods for tasks such as topic classification, sentiment analysis, or spam identification are focused, described, and analyzed in detail. Other tasks, on the other hand, like generative question answering, part of speech tagging, or machine translation are only mentioned in a non-comprehensive way. Therefore, in the context of text classification, our paper provides a comprehensive overview containing the necessary details for researchers and practitioners. For a more general perspective on NLP augmentations (including sequence-tagging, parsing, text generation, etc.), we recommend the reader to have a look at the work of Feng et al. [37], which is not as detailed in text classification as our work but presents a broader task view. In contrast to this task-driven view by Feng et al. [37], we are taking a method-oriented perspective while conducting a data-structure-driven, high-level categorization (see Figure 1). Contrary to other surveys in the field of data augmentation, we focus on setting the augmentation methods into context by comparing the conception and performance, with regard to the underlying models and application context. In this way, the listed augmentation groups contain an explanation with details on the differences within the group and a comprehensive overview of how the methods differ and which results they produce. This allows the reader to gain insights into which data augmentation technique might be most promising for the own use-case and what specifics need to be considered, while it is also possible to follow the data-structure based taxonomy. In the end, we discuss important future research directions by setting all methods into context, which can help accelerate developments in this field.

In the next section, data augmentation methods relevant in textual contexts are summarized and grouped. Generally, the methods are described in a sensible order for the specific group. In groups with many similar approaches, we summarize the most important information in tabular form. We also extract information regarding improvements. The improvement indications are intended to give a quick overview of how well a method may perform but are not in-depth informative or comparable on their own. For a more detailed perspective, the models and datasets are also displayed. This should provide a more holistic perspective, although in-depth information has to be extracted from the respective papers themselves.

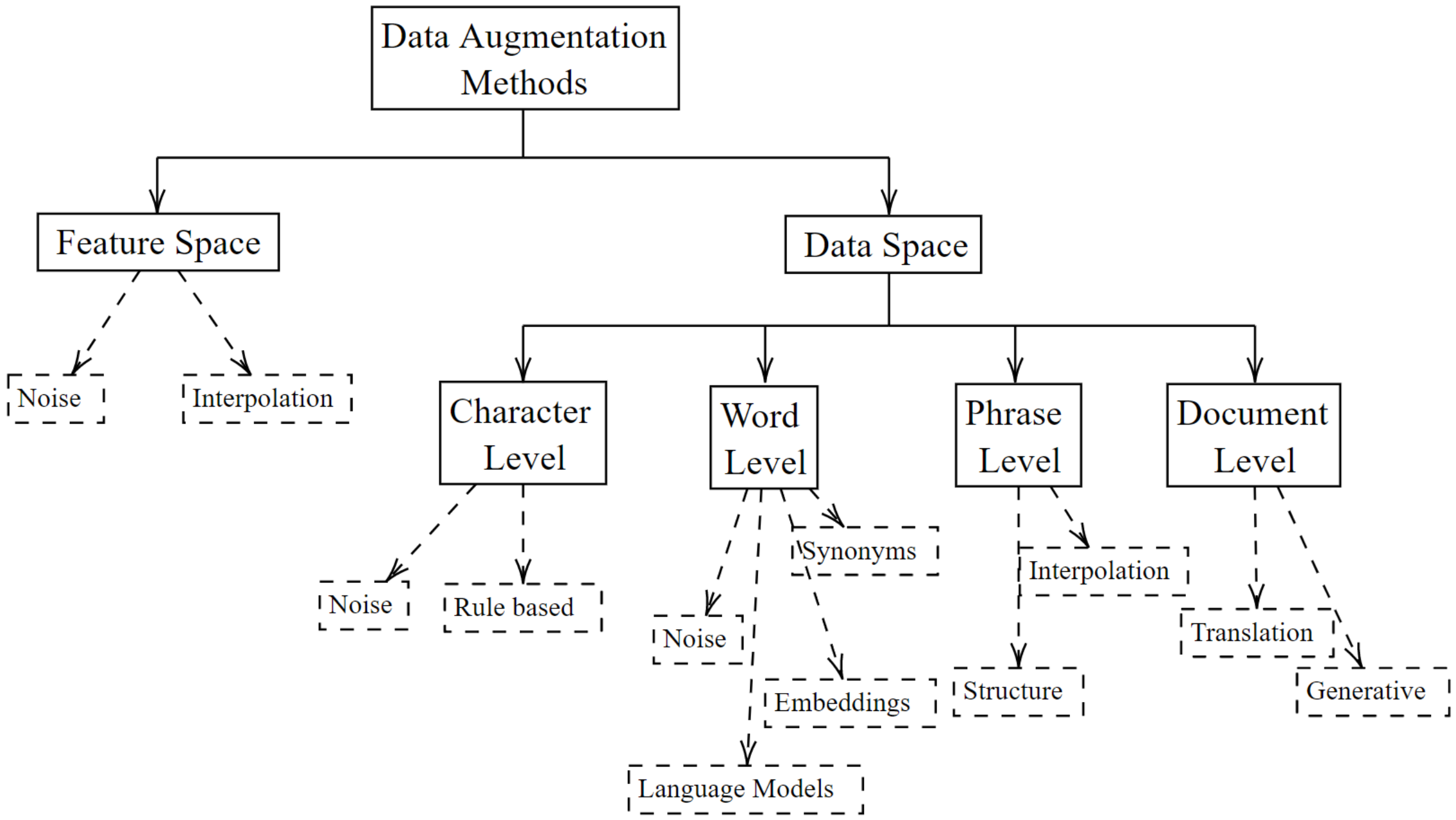

Figure 1: Taxonomy and grouping for different data augmentation methods.

### 3.1 Data Space

Augmentation in the data space deals with the transformation of the input data in its raw form, i.e., in the case of this survey, into the readable textual form of the data.

#### *3.1.1 Character Level*

##### *3.1.1.1 Noise Induction*

The addition of noise to the input data is one of the data augmentation methods with the smallest alterations, especially when applied on a character level. As explained in more detail further on, the induction of noise can also be used at the word level as well as in the feature space.

In this context, the basic idea of the method of Belinkov and Bisk [38] is to add artificial and natural noise to the training data so that, in their case, neural machine translation (NMT) models are less susceptible to adversarial examples. As artificial noise, Belinkov and Bisk [38] describe operations like the random switching of single letters ("cheese" ⟶ "cehese"), the randomization of the mid part of a word ("cheese" ⟶ "ceehse"), the complete randomization of a word ("cheese" ⟶ "eseehc"), and the replacement of one letter with a neighboring letter on the keyboard ("cheeae"). Similarly,

Feng et al. [39] randomly delete, swap, and insert characters of texts (the prompt portion) that are used for fine-tuning text generators. For this, they moreover ignore the first and last character of a word. To measure the suitability for text generators, they intrinsically measure the diversity, fluency, semantic context preservation, and sentiment consistency. The applied method is better than the baseline in every respect. These augmentations are also usable in the text classification domain. Ebrahimi et al. [21] used an existing model, trained with the initial dataset, to generate adversarial examples. They used the direct input data to flip a letter if the change increased the loss of the existing model. If a new model is trained with the additional data once again, the error rate is improved and the success of adversarial attacks is significantly mitigated. Furthermore, they compared their approach with the adversarial method from the previously mentioned work of Belinkov and Bisk [38] and the feature space method from Miyato et al. [40] (see Section 3.2.1). Based on a CharCNN-LSTM on the AG News dataset, they achieve the best improvement in accuracy by obtaining an additional 0.62%. While the method of Miyato et al. [40] improved the score by only 0.24 points, it is interesting to see that the method of Belinkov and Bisk [38] even decreased the accuracy by 0.33 points. Coulombe [9] describes the induction of weak textual sounds through the aforementioned change, deletion, and addition of letters in words and, in addition, the alteration of upper and lower case and the modification of punctuation. The highest absolute accuracy improvement by 2.5% can be seen in comparison to the best functioning baseline. However, the evaluation was performed with basic architectures and no embeddings, wherefore further studies are needed to validate the usefulness in a current setting.

Natural noise, as defined by Belinkov and Bisk [38], covers spelling mistakes that are common in the respective language, using spelling mistake databases. Each word associated with a common mistake is replaced with the misspelled word, and if there is more than one, the mistake is randomly sampled. Belinkov and Bisk [38] receive varying BLEU scores with their artificial and natural noise methods; most noise operations made the model more robust against attacks with similar operations. Most importantly, natural noise almost consistently worsens a translation model regarding the baseline. Analogous to the natural noise defined by Belinkov and Bisk [38], Coulombe [9] also adds common spelling mistakes in the textual data and achieves good improvements when added to classifiers. The best baseline (XGBoost) was improved by an additional 1.5%. With such transformations, learners are better able to deal with spelling mistakes in prospective texts, even if mistakes are not present in the original training dataset. This variant of data augmentation can for example be of interest when dealing with texts originating from social networks.

#### *3.1.1.2 Rule-based Transformations*

Coulombe [9] implements rule-based transformations through the use of regular expressions. According to him, such rules are not easy to establish, since many surface-level transformations require deeper changes to preserve the grammar, and other transformations depend on the language. Valid transformations are, amongst others, the insertion of spelling mistakes, data alterations, entity names, and abbreviations. Coulombe concretely implements the transformation of verbal short forms to their long forms and vice versa ("I am" ↔ "I'm"). In the English language, this is semantically invariant if ambiguities are preserved [9]. With this form of data augmentation Coulombe achieves very good results [9]. The best baseline model (XGBoost) additionally gained 0.5% in terms of accuracy.

### *3.1.2 Word Level*

#### *3.1.2.1 Noise Induction*

Noise induction can also be applied on the word level. For example, the method of Xie et al. [35] encompasses two noise patterns. With "unigram noising", words in the input data are replaced by another word with a certain probability. By the method of "blank noising", words get replaced with "_". By the adoption of both patterns, the authors achieved improved results in their experiments.

Li et al. [41] are using syntactic and semantic methods as well as word dropout for the generation of noise. Syntactic noise is realized via the shortening of sentences and methods such as the alteration of adjectives or the relativization of modifiers, while semantic noise is generated by the lexical substitution of word synonyms (see 3.1.2.2). In contrast to these two methods, word dropout is more clearly comparable to noise. Random input neurons or rather words get masked out during the training of the network. According to the authors, their proposed methods achieve an improvement. Especially a combination of all methods promises an improvement of up to 1.7 points in terms of accuracy.

Two of the four sub-methods of the Easy Data Augmentation (EDA) method by Wei and Zou [2], i.e., random swap and deletion, should also be mentioned as methods of noise induction. In experiments, a combination of both sub-methods led to improved performance of the used classifier. EDA is very popular in the research field and was used as a method for comparisons in the works of Qiu et al. [42], Huong and Hoang [43], Anaby-Tavor et al. [44], Kumar et al. [45], Bayer et al. [46], Feng et al. [39], Luu et al. [47], and Kashefi and Hwa [48]. Wei and Zou [2] report that for a small dataset these two sub-methods gain higher improvements than the other two sub-methods that are based on synonym replacement and insertion (see Section 3.1.2). Nevertheless, Qiu et al. [42], Anaby-Tavor et al. [44], Bayer et al. [46], and Luu et al. [47] also report some cases in which EDA as a whole data augmentation method decreases the classification score. This result can be expected, as the methods random swap and deletion are not label preserving, for example, for sentiment classification: "I did not like the movie, but the popcorn was good" ⟶random_swap⟶ "I did like the movie, but the popcorn was not good". While Wu et al. [49] also use random swap and random deletion, they propose random span deletion, where consecutive words are deleted. This technique would lead to a worse label preservation, but it is only used for language modelling with contrastive learning (see Section 3.4).

The training instances of one batch must have the same length when being fed into a neural network. For this purpose, the sequences are often zero-padded on one side. Rizos et al. [50] propose a specific noise induction method to augment the training data by shifting the instances within the confines of their padding so that the padding is not solely on one side. Evaluated by means of a hate speech detection dataset, the authors show that this method achieves additive performance gains of more than 5.8% (Macro-F1). Sun and He [51] also translate the instances by adding meaningless words either at the beginning or at the end. Unfortunately, they do not evaluate the impact of this method in isolation.

Xie et al. [52] propose a TF-IDF based replacement method in which they are replacing uninformative words of an instance with other uninformative words. As the authors are combining this technique with round trip translation (see Section 3.1.4) and unsupervised data augmentation, it is not clear to which degree it benefits the task. Similarly, Choi et al. [53] replace casual features/words that are a determining factor for the label. In the contrastive learning scheme, they mask these words to generate counterfactual examples as well as other non-casual words to generate normal augmentations. More details and results of this procedure can be found in Section 3.4.

More noise data augmentation methods related to other tasks can be found in the works of Cheng et al. [54], Li et al. [41], Wang et al. [55], Andreas [56], Guo et al. [57], Kashefi and Hwa [48], Sun and He [43], and Kurata et al. [58].

#### *3.1.2.2 Synonym Replacement*

This very popular form of data augmentation describes the paraphrasing transformation of text instances by replacing certain words with synonyms. One of the first applications of this substitution in the field of data augmentation was introduced by Kolomiyets et al. [59]. They substituted temporal expressions with potential synonyms from WordNet [60]. As the authors argue, the replacement of one original token in a sentence will mostly preserve the semantics. Based on the time expression recognition task, the authors propose replacing the headword, since temporal trigger words are usually found there. While this application, however, showed no substantial improvements, the authors also proposed a language model replacement method that was more suited for the task at hand (see Section 3.1.2.4).

In later years, many researchers experimented with word replacements based on thesauri. The works of Li et al. [41], Mosolova et al. [61], Wang et al. [62], and many more partially or primarily execute synonym substitution in this way. Differences between the studies concern the specific words that are substituted, the synonyms that come into question, and the utilization of different databases. For example, X. Zhang et al. [63] and Marivate and Sefara [64] choose the synonyms for substitution on basis of the geometric distribution by which the insertion of a rather distant synonym becomes less probable. Furthermore, several approaches exclude stop words or words with certain POS-tags from the set of words considered for replacement. Interesting is also the second sub-method of EDA by Wei and Zou [2], where synonyms are not replacing specific words, but are randomly inserted into the instance. The replacement method, synonym selection, database, and improvements of the various approaches are listed in Table 2.

Table 2: Overview of different approaches of the synonym replacement method.

| | **Synonym Database** | **Replacement Method** | **Synonym Selection** | **Model Base** | **Dataset** | **Improvements** |
|---|---|---|---|---|---|---|
| [59] | WordNet | Headword replacement | Not stated | Logistic Regression | TempEval<br>Reuters (12)<br>Wikipedia (1) | -1 (F1)<br>-0.6<br>-0.1 |
| [63], [65] | mytheas (LibreOffice)<br>WordNet-based | Randomly chosen number of words based on geometric distribution. | Randomly based on geometric distribution. | Character CNN | <br>AG News<br>DBP.<br>Yelp P.<br>Yelp F.<br>Yahoo A.<br>Amazon F.<br>Amazon P. | [63] / [65] (Acc.)<br>-0.38 / -0.57<br>+0.05 / +0.13<br>-0.03 /<br>+0.36<br>+0.22 / 0.65<br>+0.1 / 0.1<br>-0.17 / -0.17 |
| [41] | WordNet | Substitutable words are nouns, verbs, adjectives, or adverbs that are not part of a named entity.<br>Each word is replaced with a certain probability. | The remaining probability of substitution is shared among the synonyms based on a language model score. | CNN | MR<br>CR<br>Subj<br>SST<br>MR/CR<br>CR/MR | +0.8 (Acc.)<br>+1.2<br>+0.5<br>+0.1<br>0.9<br>0.3 |
| [9] | WordNet | Only adverbs and adjectives, sometimes nouns, more rarely verbs. | Most similar companion information of the synonym with the context of the chosen word. | XGBoost<br>MLP (2 hidden layer) | IMDB | +0.5 (Acc.)<br>+4.92 |
| [61] | WordNet | No pronouns, conjunctions, prepositions, and articles for replacement.<br>Choosing uniform randomly. | Uniform random | CNN with word embedding | Toxic Comment Classification | -0.09/-0.21 (AUC) |
| [62] | HIT IR-Lab Tongyici | No time words, prepositions, and mimetic words. | Chi-square statistics method | Character CNN-SVM | Hotel R.<br>Laptop R. | ~+1 (Acc.)<br>~+1 |
| cont. [62] | Cilin (Extended) (Chinese) | Chi-square statistics method. | | | Book R. | ~+0.25 |
| [64] | WordNet | Verbs, nouns, and their combination. Geometric distribution. | Geometric distribution | DNN | AG News<br>Sentiment<br>Hate Speech | ~+0.4 (Acc.)<br>~+0<br>~-0.8 |
| [66] | WordNet & Thesaurus.com | For Minibatch: Augmentation with probability,<br>POS-tag replacement, replacement of one word per sentence that maximizes loss. | Synonym that maximizes the loss. | Kim CNN | TREC | +1.2 (Acc.) |
| [2] | WordNet | No stop words.<br>Choosing n random words to be replaced (SR) or from which the synonyms are inserted at a random position (RI) | Uniform random | CNN | Classification tasks (500)<br>(2000)<br>(5000)<br>(full) | SR / RI (Acc.)<br>~+1.9 / ~+2.0<br>~+1.2 / ~+0.9<br>~+0.7 / ~+0.6<br>~+1.0 / ~+0.9 |
| [1] | WordNet | Replacement of a word based on a certain probability. | Temperature hyperparameter learned while training. | CNN | SST-5<br>SST-2<br>Subj<br>MPQA<br>RT<br>TREC | -0.6 (Acc.)<br>+0.5<br>+0<br>+0.2<br>+0.1<br>-0.4 |
| [42] | WordNet | Replacement of a word based on a certain probability. | Temperature hyperparameter learned. | TextRCNN | ICS<br>NEWS | -0.26 (Macro F1)<br>+1.63 |
| [51] | Not stated | Filtering words according to their POS-tag. Fixed or variable number of words. | Specific or variable number of synonyms. | LSTM-CNN | Tan<br>NLPCC | Results only in combination with other methods |
| [67] | WordNet | Not stated | Not stated | BERT | SST-5 (40)<br>IMDB (40)<br>TREC (40) | -0.87 (Acc.)<br>-0.87<br>+0.01 |
| [68] | WordNet | No stop words. 10% of documents randomly selected. | Not stated | M-BERT | CodiEsp-D<br>CodiEsp-P | +0.6 (F1)<br>-0.7 (F1) |
| [39] | WordNet | Keywords replaced are ordered by their RAKE score (e.g., the probability of being a keyword). | Randomly selected. Replacement only with same POS-tag. | No model (intrinsic evaluation with different metrics) | Yelp-LR (small subset of Yelp Reviews) | +0.015 (SBLEU)<br>-0.018 (UTR)<br>-0.02 (TTR)<br>-0.016 (RWords)<br>0 (SLOR)<br>-0.007 (BPRO)<br>+0.001 (SStd)<br>0 (SDiff) |
| [48] | WordNet | No stop words.<br>Uniform random replacement until 20% of the words in a sentence are changed. | Uniform random | CNN | Yelp P. | Only against other data augmentation methods |

Also to be emphasized is the more sophisticated integration into the learning process, as described by Jungiewicz and Pohl [66]. The authors replace words with synonyms only if the replacement with the chosen synonym maximizes the loss of the current state of the classifier model. Apart from this, there are approaches that adapt the general idea of thesauri-based replacements to perform augmentation on specific tasks, for example, in Kashefi and Hwa [48] and Feng et al. [39].

#### *3.1.2.3 Embedding Replacement*

Comparable to synonym substitution, embedding replacement methods search for words that fit as good as possible into the textual context and additionally do not alter the basic substance of the text. To achieve this, the words of the instances are translated into a latent representation space, where words of similar contexts are closer together. Accordingly, these latent spaces are based on the distributional hypothesis of distributional semantics [69], [70], which is currently mostly implemented in the form of embedding models. The selection of words that correspond to this hypothesis and are, thus, near in the representation space, implies that the newly created instances maintain a grammatic coherence, as displayed in Figure 2. Besides this advantage, Rizos et al. [50] argue that the "method encourages the downstream task to place lower emphasis on associating single words with a label and instead place higher emphasis on capturing similar sequential patterns, i.e., the context of hate speech". Benefits of this data augmentation technique in comparison to the synonym substitution method are that techniques based on the distributional hypothesis are more comprehensive and the context of texts is considered. This means that substitutions are not limited by a database, like WordNet, and that grammatically more correct sentences can be generated [71]. Furthermore, the general form of this approach can be beneficial for languages which have no access to a large thesauri but a lot of general text resources, on the basis of which the self-supervised embedding models can be easily trained [9].

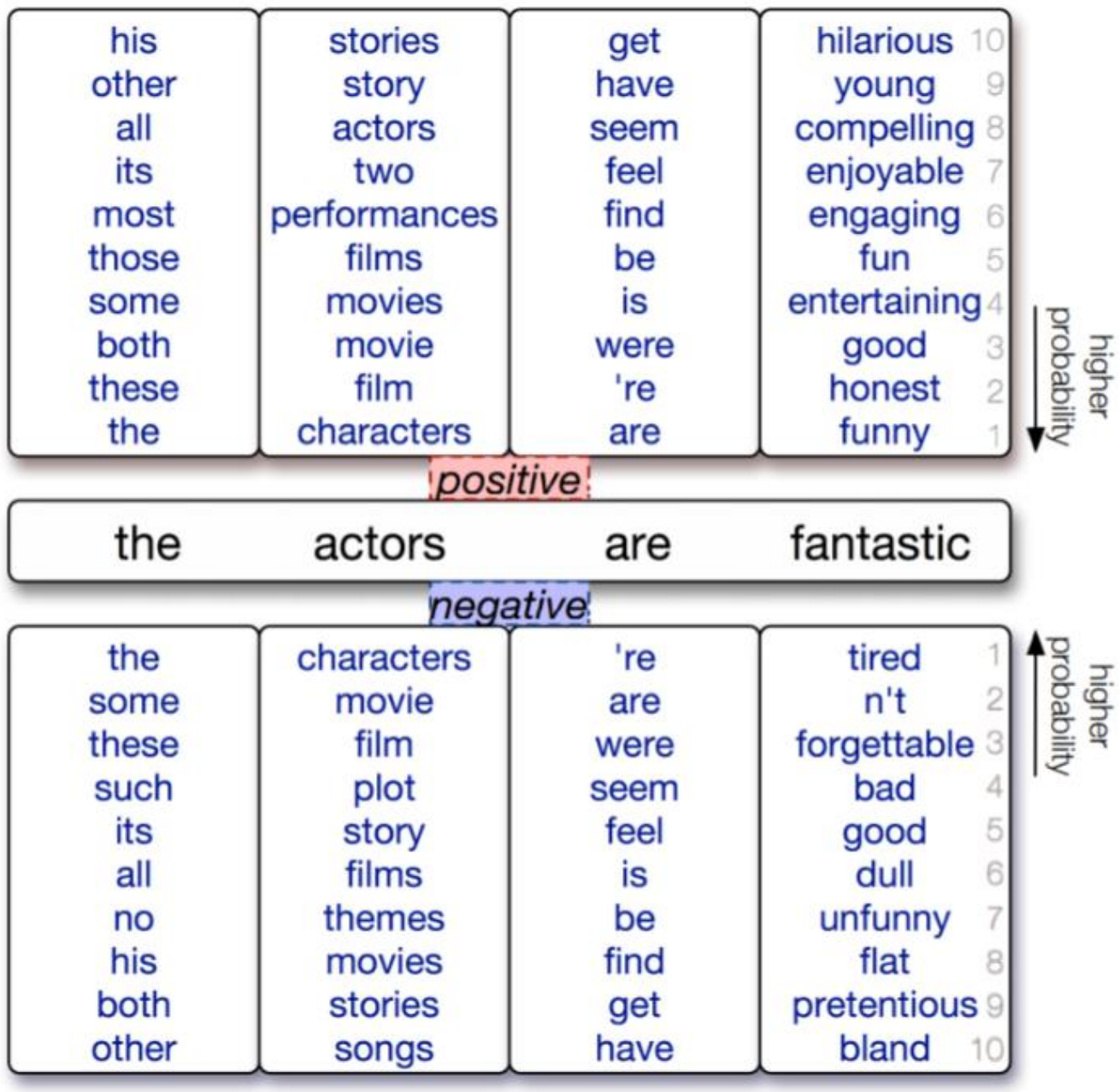

Figure 2: Example of a sentence with predicted words that can be used to replace a word in the sentence [1].

Wang and Yang [72] use this kind of augmentation to better classify annoying tweets. They utilize k-nearest-neighbor to identify the most suitable embeddings as a substitution of the training data words. Compared to the baseline, they achieve an additive improvement of up to 2.4 points in the F1-Score with logistic regression. Marivate and Sefara [64], Rizos et al. [50], Huong and Hoang [43], and others utilize the embedding replacement in very similar ways. The greatest differences in terms of the method exist in the selection of words to be replaced (e.g., POS-tag based) and the selection of the replacing words based on the embeddings. An overview of the differences can be found in Table 3.

Table 3: Overview of different approaches of the embedding replacement method.

| | Replacement Selection | Embedding Selection | Model Base | Dataset | Embedding Model | Improvements |
|---|---|---|---|---|---|---|
| [72] | Not stated | K-nearest-neighbor and cosine similarity | Logistic regression | Petpeeve dataset | UrbanDictionary W2V<br>Twitter W2V<br>GoogleNews W2V | +0.3 (F1)<br>+1.7<br>+2.4 |
| [64] | Random | Random with probability proportional to cosine similarity. | DNN | AG News<br>Sentiment<br>Hate Speech | Wikipedia W2V<br>Wikipedia W2V<br>GloVe Twitter | ~0 (Acc.)<br>~+0.5<br>~-0.3 |
| [50] | Every word | Cosine similarity threshold + POS-tag matching | CNN+LSTM/GRU | HON<br><br><br><br>RSN-1<br>RSN-2 | Word2Vec Hate Speech<br>FastText Wikipedia<br>GoogleNews W2V<br>GloVe Common Crawl<br>GloVe Common Crawl<br>GloVe Common Crawl | -22.7 (Macro F1)<br>+1.0<br>-3.3<br>+0.3<br>-0.2<br>0 |
| [51] | 1. Method: Filtering words according to their POS-tag. Selecting a fixed or variable number of words.<br>2. Method: Replacing adverbial phrases (Chinese related). | Own similarity measure and specific or variable number of replacements | LSTM-CNN | Tan | W2V self-pretrained | Results only in combination with other word level augmentation methods |
| [41] | Substitutable words are nouns, verbs, adjectives, or adverbs that are not part of a named entity. Each word is replaced with a certain probability. | Embeddings are found with counter-fitting. Candidates are replaced with a probability. The remaining probability is shared among the embeddings based on a LM | CNN | MR<br>CR<br>Subj<br>SST<br>MR/CR<br>CR/MR | GoogleNews W2V | -0.6/-4.2 (Acc)<br>+0.1/-3.7<br>+0.2/-1.4<br>-0.4/-4.2<br>+1.9/+0.4<br>+0.1/-3.0 |
| [43] | Not stated | Cosine similarity | Random Forest, Naïve Bayes, SVM | Vietnamese comments | W2V Vietnamese | Results only in combination |
| [22] | Random sampling with probabilities proportional to the neighbors each word has within the counter-fitted embedding space + exclusion of common articles and prepositions. | 1. K-nearest-neighbors with Euclidean distance + counter-fitting method.<br>2. Google LM to filter out words.<br>3. Selection of the word that will maximize the target label prediction probability. | LSTM | IMDB | GloVe | Adversarial training: No improvements but safer model |
| [73] | Only for multi-piece words. Random probability for replacement. | Random embedding of the k-nearest-neighbor | Small transformer model | Various GLUE tasks | GloVe | No augmentation baseline comparisons |
| [74] | No stop words or symbolic and numerical data | Cosine similarity threshold of 0.97 | Manhattan LSTM model | Thai text similarity task | Thai2fit (Thai language) | +1.71 |

A major factor for poor results is that the embedding replacement does not necessarily guarantee the preservation of the contextual meaning of the instances. This, in turn, could lead to distortions of the label; e.g., "the movie was fantastic" and "the movie was horrible" are valid transformations but the sentiment is the opposite. A way to address this issue is the use of the counter-fitting method of Mrkšić et al. [75] for synonym embedding substitution, as carried out by Li et al. [41]. Counter-fitting is an approach that depicts word embeddings on the basis of a target function in a way that similarities between synonyms are rewarded and similarities between antonyms are sanctioned [75]. Li et al. [41] extend this approach by selecting the most fitting words with a higher possibility for the replacement. This is done by leveraging a language model that can give an indication on how well a given word fits into a sequence. However, the authors achieve rather mixed results with this method. The counter-fitting method offers considerably less replacement possibilities, since embeddings have to be trained on the downstream task, leading to a smaller coverage of their corpora words. Alzantot et al. [22] use this method in combination with a language model filtering in their adversarial example generator. They extend the approach by only incorporating the words that are maximizing the target label prediction probability (label preservation) of an already trained classifier. The authors report no improvements in terms of the task testing set, but they show that the model is safer regarding adversarial attacks. Embedding replacement methods are moreover used in specific task-dependent ways, such as by Kashefi and Hwa [48].

#### *3.1.2.4 Replacement by Language Models*

Language models represent language by predicting subsequent or missing words on the basis of the previous or surrounding context (classical and respectively masked language modelling). In this way, the models can, for example, be used to filter unfitting words, as already discussed in Section 3.1.2.3 in relation to the work of Alzantot et al. [22]. The authors generate similar words with GloVe embeddings and the counter-fitting method and utilize a language model to choose only words with a high probability of fit. In contrast to embedding replacements by word embeddings that take into account a global context, language models enable a more localized replacement [64]. Utilizing the subsequent word prediction, language models can also be used as the main augmentation method. Kobayashi [1] is, for example, using an LSTM language model to identify substitution words. However, language models do not only substitute words with similar meaning, but also with words that fit the context in principle [1]. This trait is encompassed with a greater risk of label distortion. To prevent the attachment of wrong labels to the new training data due to changed semantics, Kobayashi [1] modifies the language model so that it allows the integration of the label in the model for the word prediction (label-conditional language model). Inspired by this approach, Wu et al. [76] alter the architecture of the language model BERT [77] in a way that it is label conditional (c-BERT). In an evaluation with different tasks the authors showed that in comparison to Kobayashi [1] and other approaches they were able to considerably increase the performance of a classifier (see Table 4). However, the c-BERT approach also has the disadvantage that the language model is fixed when applied, and in the case of low-data regimes, the augmentation might no longer be label preserving [67]. For this reason, Hu et al. [67] include the c-BERT method in a reinforcement learning scheme, which learns the task in a normal supervised fashion but is also able to simultaneously fine-tune the c-BERT LM. With this adaption, the authors significantly outperform the original c-BERT approach in a low-data regime setting. The results can be found in Table 4 together with the results of Anaby-Tavor et al. [44], who evaluated c-BERT as comparison, and Qu et al. [78], who employed the c-BERT model with supervised consistency training (see 3.4) on the MLNI-m task.

Table 4: Evaluation results of the state-of-the-art language substitution method c-BERT.

| Publication | Method | Dataset | Improvements (Accuracy) |
|---|---|---|---|
| [76] | c-BERT | SST-5 | +0.8 (CNN)/+1.3 (RNN) |
| | | SST-2 | +0.2 (CNN)/ +0.5 (RNN) |
| | | Subj | +0.5 (CNN)/ +0.4 (RNN) |
| | | MPQA | +0.5 (CNN)/ +0.7 (RNN) |
| | | RT | +0.8 (CNN)/ +0.6 (RNN) |
| | | TREC | +0.8 (CNN)/ +0.2 (RNN) |
| [78] | c-BERT with consistency training | MLNI-m | +0.4 (RoBERTa-Base) |
| [44] | c-BERT | ATIS | -1.9 (BERT) / -0.8 (SVM) / -5.8 (LSTM) |
| | | TREC | +1.1 (BERT) / +1.1 (SVM) / +6.5 (LSTM) |
| | | WVA | +0.2 (BERT) / 0.5 (SVM) / +2.4 (LSTM) |
| [67] | c-BERT integrated in reinforcement learning scheme | SST-5 (42) | +1.17 (BERT) / +2.19 (normal c-BERT) |
| | | IMDB (45) | +1.97 (BERT) / +1.97 (normal c-BERT) |
| | | TREC (45) | +0.73 (BERT) / +0.87 (normal c-BERT) |
| [73] | c-BERT and embedding substitution for multiple-pieces words | MNLI-m | +2.3 (TinyBERT) |
| | | MNLI-mm | +1.9 (TinyBERT) |
| | | MRPC | +3.4 (TinyBERT) |
| | | CoLA | +21.0 (TinyBERT) |

Jiao et al. [73] apply the already improved method by Wu et al. [76] and further adjust it in their work on TinyBERT. In doing so, the scholars reflect on the fact that the quality of the data generated with BERT is poor if many multiple-pieces words are included. To mitigate this problem, they propose to perform a embedding substitution on the base of GloVe embeddings [79] for such words. Further language model augmentations for different tasks are proposed by Gao et al. [80], Ratner et al. [81], Fadaee et al. [82], and Kashefi and Hwa [48].

### 3.1.3 *Phrase and Sentence Level*

#### 3.1.3.1 *Structure-based Transformation*

Structure-based approaches of data augmentation may utilize certain features or components of a structure to generate modified texts. Such structures can be based on grammatical formalities, for example, dependency and constituent grammars or POS-tags. Such approaches are therefore more limited to certain languages or tasks. Şahin and Steedman [83] are, for example, concerned with the augmentation of datasets from low resource languages for POS-tagging. By the method of "cropping", sentences are shortened by putting the focus on subjects and objects. With the "rotation" technique, flexible fragments are moved. The authors state that this method is dependent on certain grammatical sentence structures in different languages and probably only generates noise in the English language. Both methods are well suited for a multitude of low resource languages. They were also tested by Vania et al. [84] for the augmentation of training data for dependency parsers for low-resource data.

Feng et al. [85] propose a method for changing the semantics of a text while trying to preserve the fluency and sentiment. Given a set of phrases (replacement entities) to every instance, the so-called Semantic Text Exchange method first identifies phrases in the original text that can be replaced by a replacement entity based on the constituents. Then phrases similar to the identified phrases are replaced by a masked token. Subsequently, this is filled by an attention-based language model so that the similar words better suit the replacement entity. Feng et al. [39] adapt this approach by automatically selecting the 150 of the 200 most frequent nouns from the Semantic Text Exchange training set as replacement entity candidates and splitting their Yelp Review dataset into windows, as the method is only suitable for short texts. In an analysis with this dataset Feng et al. [39] reported that the Semantic Text Exchange method decreases fluency, diversity, and semantic content preservation.

An important work was proposed by Min et al. [86] who show that inversion (swapping the subject and object part) and passivation result in a higher generalization capability in natural language inference. In fact, considering their work in comparison with preliminary work [87]–[89] suggests that BERT is able to extract the relevant syntactic information from the instances but is unable to use this information in the task, as there are too few examples in the MNLI dataset demonstrating the necessity of syntax. Here, even a limited utilization of Min et al.'s [86] data augmentation methods already helps to mitigate this problem.

#### 3.1.3.2 *Interpolation*

In numerical analysis, interpolation is a procedure to construct new data points from existing points [90]. While the formal interpolation versions are found in the feature space section, a sensible definition of interpolation in the data space of text is difficult to construct. However, the substructure substitution ($SUB^2$) method by Shi et al. [91] is considered as such in this context due to its resemblance to the feature space methods. $SUB^2$ substitutes substructures (dependents, constituents, or POS-tag sequences) of the training examples if they have the same tagged label (for example, "a [DT] cake [NN]" in an instance can be replaced with "a [DT] dog [NN]" of another instance). The variant adapted for classification views all text spans of an instance as structures and is constrained by replacement rules that can be combined or completely left out. The replacement rules are only replacing (1) same lengths spans, (2) phrases with phrases, (3) phrases of the same constituency label, and (4) spans that come from instances with the same class label. The authors show that their methods outperform the baseline when applied to low resource tasks. Their classification variant nearly doubles the accuracy on a subsample of the SST-2 and AG News datasets. Furthermore, they achieve better results than the language model augmentation c-BERT (Section 3.1.2.4). Similarly, Kim et al. [92] propose a data augmentation method based on lexicalized probabilistic context-free grammars that extracts grammar trees from an input sentence and combines/substitutes them internally and with trees from other instances of the same class. Words are replaced with other words having the same POS-tag from the other sentences of the same class and WordNet synonyms. In this way, they can achieve a considerable performance improvement when applied in a few-shot, semi-supervised learning environment.

### *3.1.4 Document Level*

#### *3.1.4.1 Round-trip Translation*

Round-trip translation[3] is an approach to produce paraphrases with the help of translation models. A word, phrase, sentence, or document is translated into another language (forward translation) and afterwards translated back into the source language (back translation) [93]. The rationale behind this is that translations of texts are often variable due to the complexity of natural language [9], which leads to various possibilities in the choice of terms or sentence structure. The process is depicted in Figure 3.

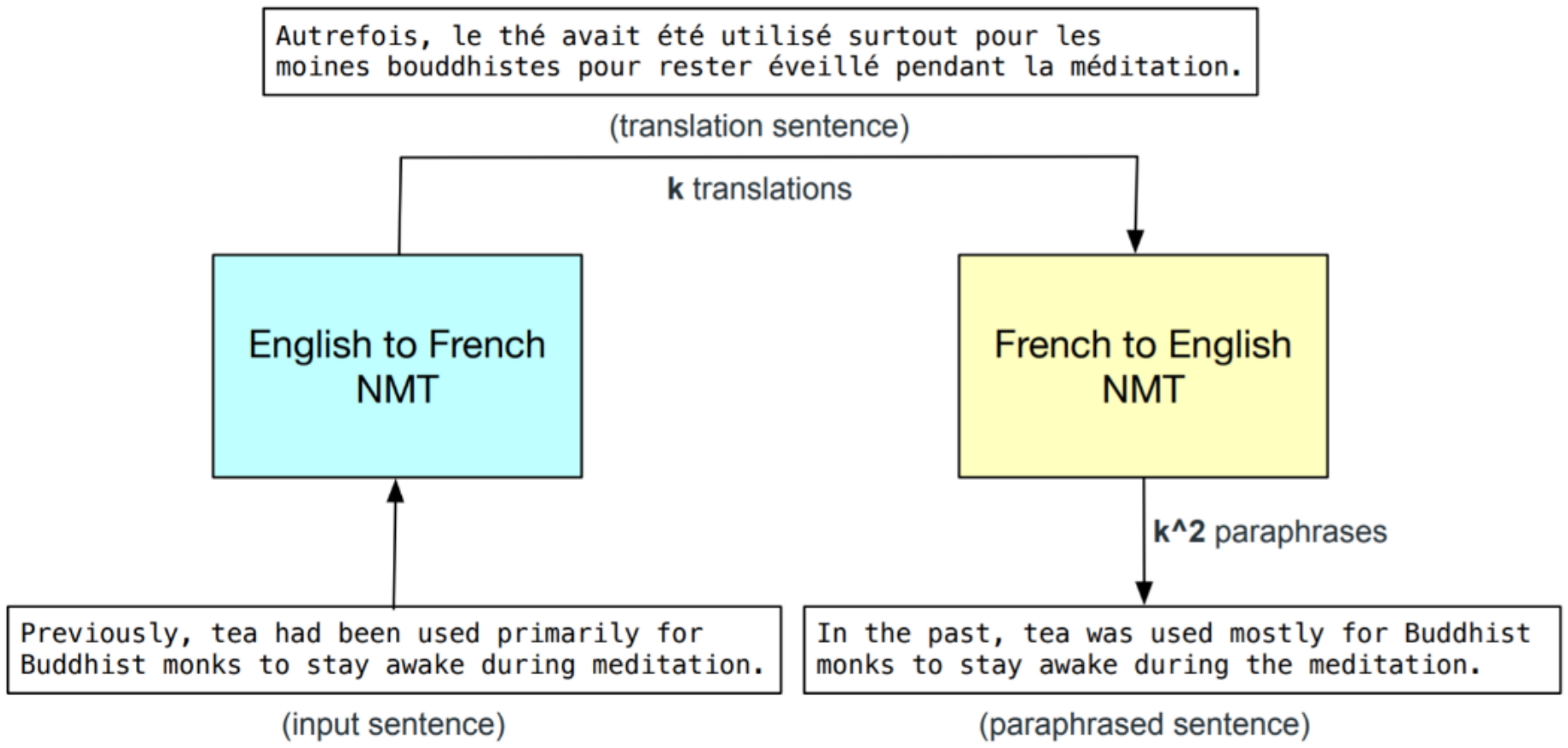

Figure 3: Round-trip translation process [94].

The approach is promising because of its good inherent label preserving and highly valuable paraphrasing capabilities. By the translation of text, the content is preserved and only stylistic features based on the traits of the author are excluded or altered [95]. Some translation systems can propose several translation options; this is hinted in Figure 2 ("k^2 paraphrases"). Yu et al. [94], Aroyehun and Gelbukh [71], Coulombe [9], Kruspe et al. [96], and others use this technique to generate artificial training data. Their works differ with regard to the used language and the subsequently applied filtering methods. These filtering methods are important, as the process of the twofold translation may be faulty [71]. Furthermore, Xie et al. [52] as well as Chen et al. [97] change the normal beam search generation strategy to random sampling with a temperature parameter to ensure a greater diversity. Details on the different round-trip translation applications are presented in Table 5.

[3] Even though Coulombe [9], Yu et al. [94], Xie et al. [52], Qu et al. [68], and others use the term backtranslation for their data augmentation works as well, these approaches are assigned to the round-trip translation approaches because they execute forward and back translation.

Table 5: Overview of the round-trip translation approaches.

| | Translation Model | Languages | Filtering | Model | Dataset | Improvements |
|---|---|---|---|---|---|---|
| [94] | Google's NMT [98] | en → fr → en | No filtering | Convolution and self-attention | SQuAD | +1.5 (EM) / +1.1 (F1) |
| [9] | Google Translate API | Not stated | Excluding identical instances. Similarity threshold based on lengths. | XGBoost<br>MLP 2 hidden layer | IMDB | +0 (Acc.)<br>+5.8 |
| [71] | Google Translate API | en → fr, es, de, hi → en | No filtering | NBSVM<br>CNN<br>LSTM<br>BiLSTM<br>CNN-LSTM<br>LSTM-CNN<br>CNN-BiLSTM<br>BiLSTM-CNN | Aggression Detection | +0.19 (Macro F1)<br>+5.31<br>+7.39<br>+5.6<br>+5.94<br>+19.45<br>+14.33<br>+6.87 |
| [96] | Google Translate | Randomly selected | No filtering | Fusion CNN | TREC Incident Streams track | ~-1.2 (F1) |
| [64] | Google Translate API & Amazon translate | en → fr, de → en | "We ensured that the [...] texts carry the same meaning as the source text" | DNN | AG News<br>Hate Speech | ~+0.33 (Acc.)<br>~-2.3 |
| [52] | WMT'14 English-French translation model | en → fr → en | No filtering | Randomly initialized transformer | Yelp-5 | +1.65 (Acc.) |
| [78] | WMT19 and released in FairSeq | en → de → en | No filtering | RoBERTa | MLNI-m | +0.9 (Acc.) |
| [99]** | Translation models from Britz et al. [100] | en → de, zh → en | No filtering | BERT | MNLI<br>QNLI<br>QQP<br>RTE<br>SST-2<br>MRPC<br>CoLA<br>STS-B | +0 (Acc.)<br>+0.2 (Acc.)<br>+0.4 (Acc.)<br>+3.6 (Acc.)<br>+0.7 (Acc.)<br>+0 (F1)<br>+2.3 (Mcc)<br>+0.6 (Corr.) |
| [101]* | Not stated | Not stated | No filtering | Transformer base with consistency training | MNLI<br>QNLI<br>QQP<br>RTE<br>SST-2<br>MRPC<br>CoLA<br>STS-B | +0.9 (Acc.)<br>+0.6 (Acc.)<br>-0.2 (Acc.)<br>+5.1 (Acc.)<br>+0.7 (Acc.)<br>+2.6 (F1)<br>+1.4 (Mcc)<br>+0.4 (Corr.) |
| [102] | MarianMT | en → fr, de, es → en Chained: en → es → fr → en | Word sense disambiguation: Retaining of those in which the target word occurs exactly once (in both original and augmented instance). | MT-DNN | SemEval-2013 + SemEval-2015 + Senseval-2 + Senseval-3 | No baseline comparisons |

* Trained with consistency training

** Trained with contrastive learning

#### 3.1.4.2 Generative Methods

Generative methods are becoming increasingly interesting in recent data augmentation research. As the capabilities of language generation increased significantly, the current models are able to create very diverse texts and can thus incorporate new information. Here, Qiu et al. [42] introduce a variational autoencoder (VAE) based on a method that is used for text generation in their system. VAEs are probabilistic neural network structures that consist of an encoder network, which transforms input data into a latent representation, and of a decoder-network, which transforms the latent representations back. The authors differentiate between unconditional and conditional VAEs. With unconditional VAEs, separate text generation models are trained for all classes, whereas with conditional VAEs, label information is fed into the model as an additional input. Furthermore, they distinguish between sampling from the prior distribution, which leads to greatly diverse instances, and the posterior distribution, which produces text that is semantically closer to the training data. With the unconditional VAE and sampling from the prior distribution, they achieve the highest improvements of up to 2 F1-points (see Table 6). Malandrakis et al. [103] make similar efforts by evaluating VAEs for augmentation. While their objective is more narrowed, as they are interested in natural language understanding with limited resources, they analyze a broader variety of VAE augmentation variants. They also propose augmentation by conditional and unconditional VAEs with sampling from the posterior or prior distribution. Furthermore, they test two different learning objectives, where in the first the VAEs are used to reconstruct the input and in the second the VAEs take an instance of a particular class and try to construct another instance from that class. They also experiment with the addition of a discriminator network to the VAE that predicts the respective class from which an output appears to be. In intrinsic and extrinsic evaluations, the conditional VAEs with the reconstruction task are best performing. The discriminator variant achieves poor results, which stem from the little amount of available training data for the many different classes. Contrary to the improvements of Qiu et al. [42], the CVAEs outperform the VAE generation. An excerpt of the extrinsic evaluation is given in Table 6. However, it must be considered that the task at hand is very specific.

VEAs are also a main component of the NeuralEditor proposed by Guu et al. [104] that generates new texts based on edition vectors. For the training of the generative model, they take pairs of instances x' and x in the training data that are lexically similar, encode the differences of them and noise into an edition vector z, and try to generate x based on x' and z. It should be noted that the lexical similarity is just a rough approximation of semantic similarity. This represents a potential source of error, as, e. g., instances could be negated which in turn weakens the label preservation capabilities. However, this suffices the purposes of the authors, as they only use the method for language modeling. Specifically, in this domain, they report improvements in terms of generation quality and perplexity. Raille et al. [105] propose Edit-transformer, which is an adaptation of the NeuralEditor with the additional ability to function cross-domain, so that the learned edits of a large dataset can be transferred to a smaller dataset. Besides the improvements in speed and language modeling, they also apply their method on three different classification tasks. The results are shown in Table 6.

Rizos et al. [50] create an RNN that, depending on a specific class, learns language modelling to generate training data thereafter. The class specific RNN for augmentation is primed with a random start word from the class specific training data. However, the authors state that this method produces the poorest results compared to embedding substitution and noise generation. In a similar sense, Ollagnier and Williams [68] perform text generation using a language model (LSTM-CNN). In contrast, they split each document in a minibatch into sentences, then generate new sentences for 30% of them and utilize 30% of the beginning of a given sentence as prompt.

Sun and He [51] use the seqGAN architecture [106] to generate artificial data on basis of a GAN. Comparable to computer vision, seqGAN consists of a generator network creating texts and a discriminator network examining the authenticity of the generated texts next to the real instances. As the discriminator network can only prove the authenticity after a sequence of words and thus gives delayed feedback to the generator, the generator network is trained as a reinforcement learning agent. Utilizing the method as a data augmentation technique, the authors only receive minor improvements of classification quality. Partially inspired by SeqGAN, Li et al. [107] propose CS-GAN, which consists of a GAN, RNN, and reinforcement learning component for sentence generation. The model receives the information about the label as a prior for the generator, which is implemented with by the RNN and RL components, which is then required by the discriminator to generate meaningful sentences. Subsequently, a classifier forces the output of sentences to fit the label. The results are listed in Table 6.

Wang and Lillis [108], Anaby-Tavor et al. [44], Abonizio and Junior [109], Bayer et al. [46], Claveau et al. [110], and Liu et al. [111] use the GPT-2 model of Radford et al. [112], which achieves very good results in text generation, to create new complete instances. Concerning the adoption of the method, Wang and Lillis [108] only describe that they use rare instances as dependent examples for the generation. Anaby-Tavor et al. [44], on the other hand, develop a method that increases the safety with regard to label preservation. In a first step, they further train the GPT-2 model with training data of a certain task (fine-tuning). In the process, they concatenate the respective label to every instance in order to facilitate the generation of new data for the respective class. Finally, a classifier determines which generated instances can actually be assigned to the class stated. The authors manage to achieve significant improvements in the classification of sentences. They show that their method outperforms conditional VAEs (unfortunately no sampling technique is described) and even EDA (Section 3.1.1.1) and c-BERT (Section 3.1.2.4) when applied to a severe low data regime. The results of their LAMBADA approach and CVEA implementation are given in Table 6. Abonizio and Junior [109] try to improve this approach by concatenating three random samples as a prompt for the generation. Furthermore, they are using DistilGPT2 by Sanh et al. [113], which is substantially faster and smaller. As can be seen in Table 6, the method consistently outperforms the baseline. While LAMBADA and PREDATOR are only applicable to short texts as instances, Bayer et al. [46] design a GPT-2 based approach to augment short as well as long text tasks. In this way, very high label preservation and diversity is to be achieved by fine-tuning the language model on the class specific data, generating data prompted with specialized training data tokens, and a filtering method based on document embeddings. They can achieve high improvements for constructed and real-world low data regimes. However, they also discuss limitations of their method and useful applications in terms of specific datasets and tasks. The results can also be seen in Table 6. Similarly, Claveau et al. [110] fine-tune the GPT-2 model using the class-specific data and input a random word from the original texts for generation. Afterwards a classifier is applied to filter the generated data instances. They evaluate their approach using English and French datasets (see Table 6). Liu et al. [111] use a reinforcement learning component after the softmax prediction of the GPT-2 model to predict the tokens depending on the class for which the instance is to be generated. The authors tested their method with various model architectures. It consistently improved all of them in all tasks, especially the larger pre-trained models, like BERT and XLNet. The results for XLNet are shown in Table 6. Yoo et al.

[114] are among the first authors using the much larger GPT-3 model by Brown et al. [115] for data augmentation, which has considerably better generation capabilities. Such large language models are expensive and hard to be fine-tuned on the training data, which is why their augmentation method GPT3Mix selects some examples from the dataset and incorporates them with the label into sensible prompts for the model to be conditioned on. The newly created instances are then extracted from the generated text and a pseudo-label probability is calculated with the GPT model. As shown in Table 6, the method achieves outstanding performance increases on scarce and one full dataset. The authors further demonstrate that their method is superior to other augmentation methods, such as EDA [2], round-trip-translation and Tmix [97], and that the performance increases if larger models for the classification are used. Nevertheless, given the size of the GPT-3 network and the correspondingly large training dataset, it might even be able to replicate some of the training (or even test) instances that were left out in the creation of a scarce dataset.

In the generative method, proposed by Lee et al. [116], a first step is to subdivide the data into slices (informed by or based on the labels). Then, a generative model is trained on these slices to predict an instance in the slice based on a subsample of instances in that slice. This model is subsequently used to generate new data for underrepresented slices by priming it with instances from it. This way, the authors gain several improvements in text classification, intent classification, and relation extraction tasks with state-of-the-art results for the latter two. Furthermore, Ding et al. [117] and Chang et al. [118] propose methods using generative models for tasks other than text classification.

Table 6: Overview of text generation methods.

| **Publication** | **Method** | **Model** | **Dataset** | **Improvements** |
|---|---|---|---|---|
| [42] | VAE | Ensemble of BiLSTM, TextCNN, TextRCNN, and FastText with XGBoost as top-level classifier | ICS (Zh) | +0.04 (F1) |
| | | | News Category Dataset (EN) | +2.02 |
| | CVAE + prior sampling | | ICS (Zh) | -0.13 |
| | | | News Category Dataset (EN) | +1.55 |
| | CVAE + posterior sampling | | ICS (Zh) | -0.06 |
| | | | News Category Dataset (EN) | +1.88 |
| [103] | VAE | BiLSTM | Movie | +4.0 (Macro F1) |
| | | | Movie + Live Entertainment | -0.5 |
| | CVAE + prior sampling | | Movie | +5.9 |
| | | | Movie + Live Entertainment | +1.7 |
| | CVAE + posterior sampling | | Movie | +5.6 |
| | | | Movie + Live Entertainment | +0.6 |
| [119] | CVAE | BERT | SNIPS (few shot) | +8.00 |
| | | | SNIPS | +0.06 (Acc.) |
| | | | FBDialog (few shot) | +7.42 |
| | | | FBDialog | +0.0 |
| [105] | Transformer-based sentence editor | CNN | Subj (20%) | +1.71 (Acc.) |
| | | CNN | Subj (100%) | +1.62 |
| | | CNN | SST-2 (20%) | +0.87 |
| | | CNN | SST-2 (100%) | -0.84 |
| | | LSTM | Amazon Reviews (1%) | +1.12 |
| | | LSTM | Amazon Reviews (4%) | +0.41 |
| [50] | RNN LM with random start word priming | CNN+LSTM + GloVe++ | HON | -1.8 (Micro-F1) |
| | | | RSN-1 | +8.2 |
| | | | RSN-2 | -7.4 |
| [68] | CNN-LSTM LM priming with 30% of a sentence | CNN-LSTM | CodiEsp-P | +3.1 (F1) |
| [51] | seqGAN | LSTM + pretrained embeddings<br>CNN + pretrained embeddings<br>LSCNN + pretrained embeddings | Tan's task | +1.06 (F1)<br>+0.9<br>+0.8 |
| [107] | CS-GAN (GAN, RNN and reinforcement learning) | CNN | Amazon-5000<br>Amazon-30000<br>Emotion-15000<br>NEWS-15000 | +1.6 (Acc.)<br>-0.21<br>+0.77<br>+2.25 |
| [108] | GPT-2 for rarer instances without filtering | Logistic regression/biLSTM/Bi-attentive classification + ELMo + GloVe | Alerting Information Feed Prioritization | No comparative results |
| [44] | CVAE | BERT | ATIS (5)<br>TREC (5)<br>WVA (5) | +7.3 (Acc.)<br>+0.8<br>-1.8 |
| [44] | GPT-2 generation and classifier filtering | BERT | ATIS (5)<br>ATIS (20)<br>ATIS (50)<br>ATIS (100)<br>TREC (5)<br>WVA (5) | +22.4 (Acc.)<br>~0<br>~+2.0<br>~+0.5<br>+4.0<br>+1.4 |
| [109] | DistilGPT2 generation and classifier filtering | BERT<br>CNN<br>BERT | AG-NEWS<br>CyberTrolls<br>SST-2 | +0.61 (Acc.)<br>+0.45<br>+1.63 |
| [46] | GPT-2 with conditional fine-tuning, special prompting, and embedding filtering | ULMFit | SST-2 (100)<br>SST-2 (700)<br>Layoff<br>Management Change<br>Mergers & Acquisitions<br>Flood<br>Wildfire<br>Boston Bombings<br>Bohol Earthquake<br>West Texas Explosions<br>Dublin<br>New York | +15.53 (Acc.)<br>-0.19 (Acc.)<br>+4.84 (F1)<br>+3.42 (F1)<br>+1.42 (F1)<br>+0.25 (F1)<br>+0.44 (F1)<br>+2.44 (F1)<br>+2.05 (F1)<br>+3.81 (F1)<br>−2.54 (F1)<br>+0.44 (F1) |
| [110] | GPT-2 with conditional fine-tuning, special prompting, and classifier filtering | RoBERTa<br>FlauBERT | MediaEval<br>CLS-FR | +0.55 (micro-F1)<br>+0.57 |
| [111] | GPT-2 with a reinforcement learning component for class conditional generation. | XLNet | Offense Detection (20%)<br>Offense Detection (40%)<br>Sentiment Analysis (20%)<br>Sentiment Analysis (40%)<br>Irony Classification (20%)<br>Irony Classification (40%) | +1.3 (F1)<br>+4.3<br>+1.2<br>+1.4<br>+1.0<br>+2.3 |
| [114] | GPT-3 with prompt-based generation and pseudo-labeling | BERT (base) | COLA (0.1%, 0.3%, 1.0%) | +7.9, 3.2, -2.4 |
| | | | TREC6 (0.1%, 0.3%, 1.0%) | +15.6, 17.1, -6.5 |
| | | | CR (0.1%, 0.3%, 1.0%) | +11.0, 17.3, 8.9 |
| | | | SUBJ (0.1%, 0.3%, 1.0%) | +1.3, -1.8, -1.2 |
| | | | MPQA (0.1%, 0.3%, 1.0%) | +12.9, 13.4, 3.8 |
| | | | RT20 (0.1%, 0.3%, 1.0%) | +6.2, 13.6, 17.5 |
| | | | SST-2 (0.1%, 0.3%, 1%, full) | +20.9, 19.3, 5.7, 2.9 |
| | | BERT (large) | SST-2 (0.1%, 0.3%, 1.0%) | +23.7, 14.6, 3.0 |

### 3.2 Feature Space

Data augmentation in the feature space is concerned with the transformation of the feature representations of the input.

#### 3.2.1 *Noise Induction*

As in the data space, noise can also be introduced in several variants in the feature space. For example, Kumar et al. [119] employ four such techniques for the ultimate goal of intent classification. One of those methods applies random multiplicative and additive noise to the feature representations, as shown in [58]. However, in contrast, they are not transforming the created representations back into the data space. Another method called Linear Delta calculates the difference between two instances and adds it to a third (all from the same class). The third method, which interpolates instances, is further elaborated in Section 3.2.2.2 (see Table 8). For their fourth method, the authors are adapting the Delta-Encoder by Schwartz et al. [120] for textual data. There, an autoencoder model learns the deltas between instance pairs of the same class, which is then utilized to generate instances of a new unseen class. In a normal testing setting, the methods only slightly improve the classification results, while in a few-shot setting all methods are highly beneficial.

Several feature space data augmentation methods stem from the adversarial training research field. As explained in the background section, the models are trained with adversarial examples, i.e., little perturbed training data instances that would change the prediction or maximize the loss. This can be formally written as follows [121]:

$$\min_{\boldsymbol{\theta}} \mathbb{E}_{(\mathbf{Z},y) \sim D} \left[ \max_{||\delta|| \leq \epsilon} L(f_{\boldsymbol{\theta}}(\mathbf{X} + \boldsymbol{\delta}), y) \right],$$

where Θ are the model parameters and $\delta$ describes the perturbation noise added to the original instances (within a norm ball). Further, $D$ is the data distribution, $y$ the label, and $L$ a loss function. The training of the network (outer minimization) can still be solved by stochastic gradient descent (SGD), while the search for the right perturbations (i.e., inner maximization) is non-concave [121]. As described by Zhu et al. [121], projected gradient descent (PGD) [122], [123] can be used to solve this. Unfortunately, several convergence steps (K) to get a good result make it computationally expensive [121]. Shafahi et al. [124] and Zhang et al. [125] propose two methods that calculate the gradient with respect to the input (for PGD) on the same backward pass as the gradient calculations regarding the network parameters during a training step. This mitigates additional calculation overhead of PGD. In detail, Free adversarial training (FreeAT) by Shafahi et al. [124] trains the same batch of training examples K times so that several adversarial updates can be performed. You Only Propagate Once (YOPO) by Zhang et al. [125] accumulates the gradients with respect to the parameters from the K steps and updates the parameters accordingly. Zhu et al. [121] also propose a method called Free Large-Batch (FreeLB), which is similar to YOPO, as it also accumulates the parameter gradients. On several tasks, this method consistently exceeds the results of the baseline and the other two methods. The results of the GLUE dataset are given in Table 7. Miyato et al. and Miyato et al. [40], [126] change the normal adversarial training rule so that no label information is needed and call it *virtual adversarial training*. Without going into exact details, virtual adversarial training regularizes the standard training loss with a KL divergence loss of the distribution of the predictions with and without perturbations, where the perturbations are chosen to maximize the KL divergence. While the virtual adversarial training method is suitable for semi-supervised learning, we are particularly interested in the supervised setting. Their method improves the supervised DBpedia topic classification task baseline classifier by 0.11 points of accuracy, leading to a + 0.03 increase in accuracy in comparison to the conventional adversarial training method. Jiang et al. [127] propose the adversarial method SMART, which relies on the virtual adversarial training regularization. They introduce the utilization of the Bregman proximal point optimization with momentum to solve the virtual adversarial training loss, which prevents the model from aggressive updates [40]. The authors show in their experiments that the method significantly improves the baseline and is also able to achieve better results than the other methods discussed in this paragraph (for an overview, see Table 7). Furthermore, they demonstrate robustness enhancement and domain adaption capabilities in several evaluation applications.

Wang et al. [128] and Liu et al. [129] developed methods for enhancing the pre-training of language models with adversarial training. Wang et al. [128] simply generate adversarial examples on the output embeddings in the softmax function of the language models. Thereby they manage to reduce the perplexity of the AWD-LSTM and QRNN models on different datasets, which leads, for example, to a reduction of 2.29 points with respect to the Penn Treebank dataset with the AWD-LSTM model. However, it is not clear how the training of bigger pre-trained language models like BERT and RoBERTa would have been influenced by this method. This is addressed in the work of Liu et al. [129] with their method called Adversarial training for large neural Language Models (ALUM), which introduces noise to the input embeddings. The authors build their system based on the virtual adversarial training by Miyato et al. [40], as they noticed that it is superior to conventional adversarial training for self-supervision. Furthermore, they found out that they can omit the Bregman proximate point method and the adversarial training proposed by Jiang et al. [127] and Shafahi et al. [124] when they are using curriculum learning, where the model is first trained with the standard objective and then with virtual adversarial training. They report promising generalization and robustness improvements with the largest transformer models. For example, RoBERTa models can be improved with the ALUM continual pretraining by + 0.7 on the MNLI task, while standard continual pretraining does not introduce further gains. The results on the GLUE dataset are given in Table 7. The authors also tested the robustness of the models with three different adversarial datasets, where ALUM achieves significant improvements in all tasks. In another evaluation setting they combine adversarial pretraining with adversarial fine-tuning. ALUM improves all evaluation scores of the standard pretrained models. This model reaches the best performances and significantly outperforms the other models in all tested tasks, e.g., with an increased accuracy of + 0.4 more than without tuning the SNLI dataset. The improvement on the MNLI task is given in Table 7.

With regard to the generative adversarial training methods of the feature space, it is also of interest to investigate how the newly created examples can be transformed into the data space to enable their inspection. This is done in the works of Liu et al. and Wan et al. [130], [131]. Wan et al. attempt to improve the classification behavior of a grammatical error correction system by training with adversarial examples. Such an example, extracted from the application of loss-increasing noise in the hidden representation of a transformer encoder, is mapped to the data space by a transformer encoder that was trained autoregressively. Then they use a similarity discriminator based on the model to filter instances that are not similar to their initial counterparts. Liu et al. [130] also use a transformer autoencoder architecture to generate data space instances. In contrast to the work of Wan et al. [131], they generate the noisy instances from the input embeddings, subsequently filter instances based on unigram word overlap, and try to improve machine question generation and question answering tasks. Both methods significantly improve the baselines and other methods.

Given the constraint that adversarial training can be computationally expensive, Shen et al. [101] propose three simple and efficient data augmentation methods of the feature space (see Figure 4). Token cutoff sets the entire embedding of a single word to 0, while the feature cutoff sets one embedding dimension of each word in the input to 0. The third method, span cutoff, employs token cutoff across a coherent span of words. With each method, several different, slightly modified instances can be created, which the authors see as different perspectives/views that can be integrated in a multi-view learning fashion through consistency training. This means that the model should predict similar outputs across different views (details can be found in Section 3.4). The authors evaluate their model on the GLUE task and compare it with adversarial training algorithms as well as round-trip translation. In three out of eight tasks, an improvement over all other methods could be achieved (see Table 7). They extend the cutoff strategies to work with language generation, and thereby significantly outperform the baseline as well as the adversarial training method of Wang et al. [128].

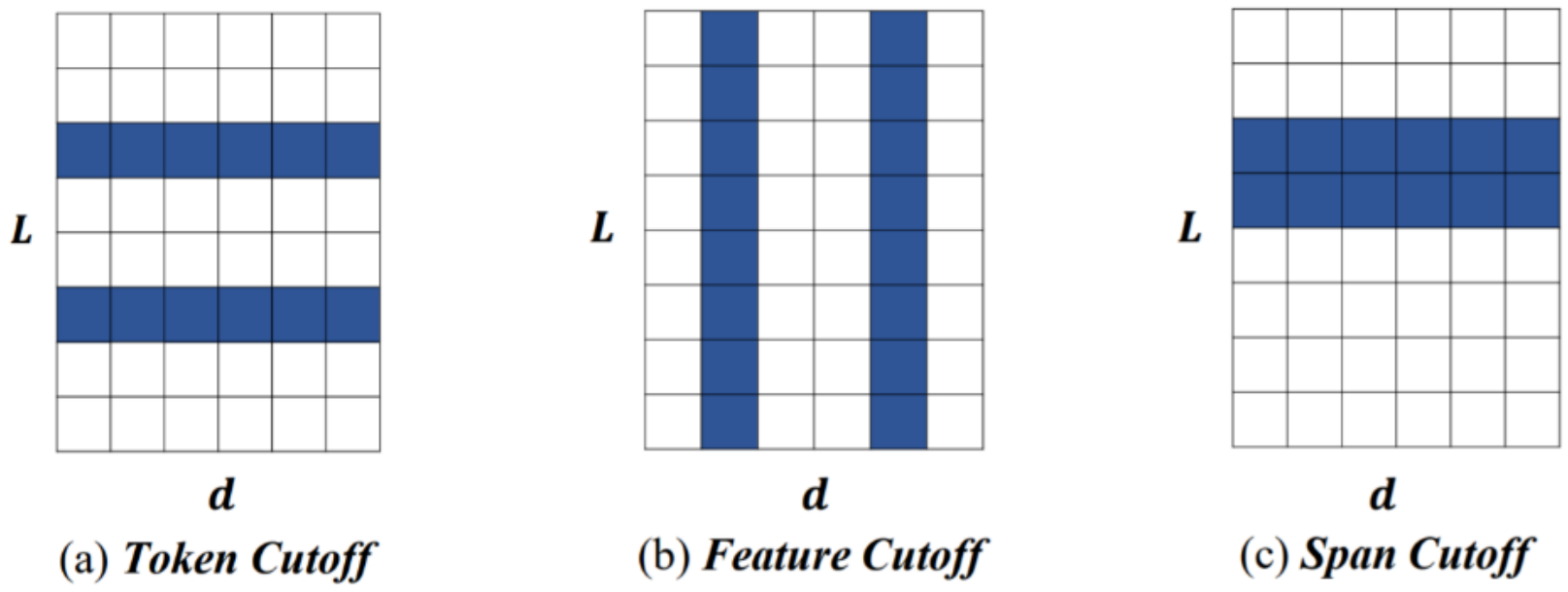

Figure 4: Visualization of the different cutoff methods [101].

Table 7: Comparison of different noise inducing methods on the GLUE task.

| | Model | SST-2 Acc | STS-B P/S Corr | MNLI-m/mm-Acc | QQP Acc | RTE Acc | QNLI Acc | MRPC F1 | CoLA Mcc |
|---|---|---|---|---|---|---|---|---|---|
| **Baseline** | RoBERTa-L | 96.4 | 92.4 | 90.2 | 92.2 | 86.6 | 94.7 | 90.9 | 68.0 |
| **Adversarial Training** | PGD | 96.4 | 92.4 | 90.5 | 92.5 | 87.4 | 94.9 | 90.9 | 69.7 |
| | FreeAT | 96.1 | 92.4 | 90.0 | 92.5 | 86.7 | 94.7 | 90.7 | 68.8 |
| | FreeLB | 96.7 | 92.7 | 90.6 | **92.6** | 88.1 | 95.0 | 91.4 | 71.1 |
| | ALUM* | 96.6 | 92.1 | 90.9 | 92.2 | 87.3 | 95.1 | 91.1 | 68.2 |
| | ALUM | - | - | **91.4** | - | - | - | - | - |
| | SMART | 96.9 | **92.8** | 91.1 | 92.4 | **92.0** | **95.6** | 92.1 | 70.6 |
| **Cutoff**** | Token | 96.9 | 92.5 | 91.0 | 92.3 | 90.6 | 95.3 | 93.2 | 70.0 |
| | Feature | **97.1** | 92.4 | 90.9 | 92.4 | 90.9 | 95.2 | 93.4 | 71.1 |
| | Span | 96.9 | **92.8** | 91.1 | 92.4 | 91.0 | 95.3 | **93.8** | **71.5** |

*only adversarial pre-training
**supervised consistency training

### 3.2.2 Interpolation Methods

For textual data, interpolation methods are mostly limited to the feature space since there is no intuitive way for combining two different text instances. Nevertheless, the application in the feature space seems reasonable, as the interpolation of hidden states of two sentences creates a new one containing the meaning of both original sentences [97], [132]. Besides this, from a learning-based perspective, interpolation methods have a high value for machine learning models. Possible explanations for the success of interpolation methods may stem from the balancing of classes, the smoothening of the decision border (regularization) [133], and the improvement of the representations [134].

For example, the Synthetic Minority Over-sampling Technique (SMOTE) approach in its original context was developed for the purpose of oversampling the minority class, which, as described in the background section, inherently leads to better classification performances. In fact, a balancing of a class can easily be achieved by simply copying the minority class. However, Chawla et al. [133] show that simple oversampling leads to more specific decision boundaries than applying SMOTE in the balancing of classes. Interpolation methods can smoothen the boundary, as shown in Figure 5. Smoothened and more general decision borders signify that an algorithm can generalize better and, in relation to training data, is accompanied by less overfitting. In this context, when applying interpolation methods to representations of the input data, Verma et al. [134] empirically and theoretically prove that representations are flattened with regard to the number of directions with significant variance. This is desirable since data representations capture less space, meaning that a classifier is more uncertain for randomly sampled representations and a form of compression is achieved which leads to generalization [134]–[136].

#### *3.2.2.1 SMOTE Interpolation*

SMOTE is an interpolation method of feature space representations of input data. With SMOTE, various neighbors close to a specific instance are searched within the feature space in order to be interpolated with the following formula:

$$\tilde{x} = x_i + \lambda * dist(x_i, x_j),$$

where $(x_i, y_i)$ is the source instance and $(x_j, y_i)$ is a close neighbor with the same class label. $dist(a, b)$ is a distance measure and $\lambda \in [0,1]$. Unlike mixup, only instances of the same class get interpolated. The rationale behind the calculation of neighbors with the same class labels is that the interpolations tend to be class preserving, leading to a higher safety of the technique. However, this leads to a limited novelty and diversity of the created instances.

SMOTE is rudimentarily illustrated in Figure 5. In the illustration, a binary classification problem is shown, in which a learning algorithm has learned the specific decision border. To achieve a balanced class distribution, a new instance is added to the blue class by utilizing SMOTE. This addition achieves, apart from a balancing of the dataset, an adjustment of the decision boundary. The new boundary is less specific and thus contributes to more general decisions. SMOTE in combination with textual data augmentation is applied, for instance, in the work of Wang and Lillis [108]. Unfortunately, the authors do not describe how and at which point of the network the method is applied.

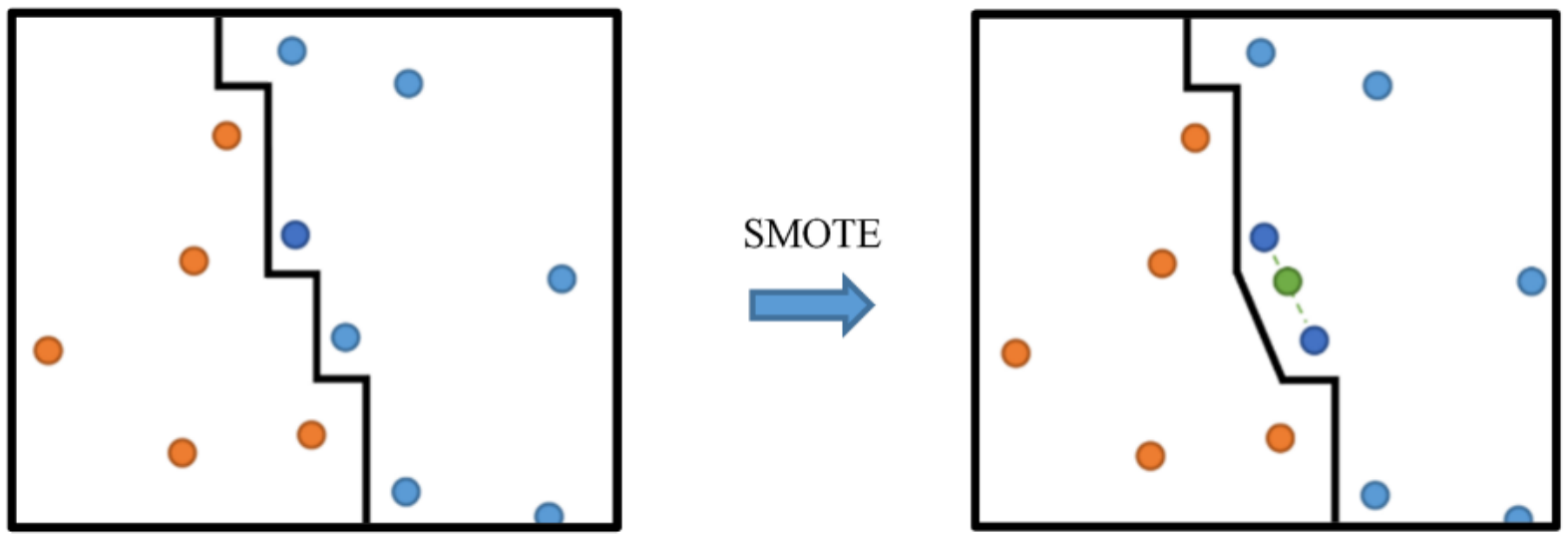

Figure 5: Illustration of the interpolation method SMOTE.

#### 3.2.2.2 *Mixup Interpolation*

Mixup by Zhang et al. [137] is an interpolation method similar to SMOTE. In the simplest adoption, the convex interpolation is implemented with the following formulas:

$$\tilde{x} = \lambda x_i + (1 - \lambda) x_j, \quad \text{whereas } x_i, x_j \text{ are input vectors}$$

$$\tilde{y} = \lambda y_i + (1 - \lambda) y_j, \quad \text{whereas } y_i, y_j \text{ are one-hot-coded labels}$$

$(x_i, y_i)$ and $(x_j, y_j)$ are sampled from the training data and $\lambda$ is either fixed in $[0,1]$ or $\lambda \sim \text{Beta}(\alpha, \alpha)$, for $\alpha \in (0, \infty)$.

Mixup is a general technique that can be applied to all kinds of equal dimensional data. However, text cannot trivially be represented in equal dimensions [138]. As a very general method, Verma et al. [134] propose the idea of applying mixup within a randomly selected hidden layer of a neural network. Despite the fact that the authors only perform the tests on image datasets, this approach paves the way for the application of mixup in many textual related tasks. The results are very promising, and for textual evaluations we advise the reader to look at the methods described below (Table 8), which oftentimes can be seen as specializations of the approach by Verma et al. [134] for textual data. Marivate and Sefara [64] state that they use mixup on representations of bag of word models, TF.IDF models, word embeddings, and language models. Unfortunately, the authors do not explicitly describe how the interpolation is performed. This raises questions about how to interpolate instances of different sizes, when, for example, word embedding vectors are used. Marivate and Sefara [64] report about 0.2, 0.4, and 0 points gain for the AG News, Sentiment 140, and Hate Speech detection task. In contrast, Qu et al. [78] describe the internal implementation of their interpolation. For the interpolation, they draw two instances from a mini-batch and linearly combine their input embedding matrices in the way described above. They improve the baseline on the MNLI-m task by an additional 0.6% in terms of accuracy. Guo et al. [139] propose two variants, wordMixup and senMixup, where the interpolation is applied in the word embedding space and on the final hidden layer of the neural network before it is passed to a softmax layer. For wordMixup the sequences have to be zero padded so that the dimensions are the same. For senMixup this is not necessary, as the hidden embeddings generated are of the same length each. The improvement results of both methods with regard to the CNN model with pretrained GloVe embeddings (trainable), which is the best baseline, is presented in Table 8. Guo [140] further advances the wordMixup approach by using a nonlinear interpolation policy. The policy is constructed to mix each dimension of the individual word embeddings in a given sentence separately. Furthermore, the labels are also interpolated nonlinearly, while they are learned adaptively based on the mixed embeddings. This way, a much larger variety of generated examples can be created. While this procedure outperforms the other two variants in most tasks, it can also have a negative effect on the classification quality, as shown in Table 8. Similar to the senMixup method, Sun et al. [138] apply mixup to the output of transformer models. Furthermore, they only activate mixup in the last half of the training epochs to learn good representations first. The improvements on the GLUE benchmark are listed in Table 8. In a very similar way, Chen et al. [97] propose TMix, which is also able to interpolate the hidden representations of an encoder. Indeed, TMix is able to interpolate at every layer of the encoder, similar to Verma et al. [134]. Based on the work of Jawahar et al. [141], who analyzed the types of information learned in different layers of BERT, the authors narrowed down their approach and opted for 7, 9, and 12 as interpolation layers as they contain the syntactic and semantic information. The improvements of TMix are also shown in Table 8.

Table 8: Overview of different approaches of the replacement method "mixup interpolation".

| Method | Technique for textual application | Model | Datasets | Improvements |
|---|---|---|---|---|
| mixup by Marivate and Sefara [64] | Not stated | DNN | AG News | +0.2 (Acc.) |
| | | | Sentiment 140 | +0.4 |
| | | | Hate Speech | +0 |
| [119] | Interpolation of the BERT CLS output | BERT-base-english-uncased | SNIPS (few shot)<br>SNIPS<br>FBDialog (few shot)<br>FBDialog | +8.36 (Acc.)<br>+0.0<br>+7.92<br>+0.08 |
| [78] | Interpolation of the embedding matrices | RoBERTa-base | MNLI-m | +0.6 (Acc.) |
| wordMixup by Guo et al. [139] | Interpolation of zero-padded word embeddings | CNN | Trec<br>SST-1<br>SST-2<br>Subj<br>MR | +1.6 (Acc.)<br>+1.9<br>+0.2<br>+0.3<br>+1.5 |
| senMixup by Guo et al. [139] | Interpolation on the final hidden layer | CNN | Trec<br>SST-1<br>SST-2<br>Subj<br>MR | +1.2 (Acc.)<br>+2.3<br>+0.3<br>+0.5<br>+0.8 |
| Nonlinear Mixup by Guo [140] | Nonlinear interpolation of padded word embeddings | CNN | Trec<br>SST-1<br>SST-2<br>Subj<br>MR | +2.6 (Acc.)<br>+3.0<br>+2.3<br>-0.5<br>+3.6 |
| Mixup-Transformer by Sun et al. [138] | Interpolation after last layer of the transformer | BERT-large | CoLA<br>SST-2<br>MRPC<br>STS-B<br>QQP<br>MNLI-mm<br>QNLI<br>RTE | +2.68 (Corr.)<br>+0.81 (Acc.)<br>+1.72 (Acc.)<br>+0.89 (Corr.)<br>+0.42 (Acc.)<br>-0.01 (Acc.)<br>+0.13 (Acc.)<br>+0.37(Acc.) |
| TMix by Chen et al. [97] | Interpolation of the m-th BERT layer (7, 9, and 12 randomly chosen per batch) | BERT-base-uncased + average pooling + two-layer MLP | AG News (10)<br>AG News (2500)<br>DBPedia (10)<br>DBPedia (2500)<br>Yahoo! (10)<br>Yahoo! (2500)<br>IMDB (10)<br>IMDB (2500) | +4.6 (Acc.)<br>+0.2<br>+1.6<br>+0.0<br>+2.4<br>+0.3<br>+1.8<br>+0.5 |
| TMix evaluated by [114] | Interpolation of the m-th BERT layer (7, 9, and 12 randomly chosen per batch) | BERT-base | SST-2 (0.1, 0.3, 1.0%)<br>COLA (0.1, 0.3, 1.0%)<br>TREC6 (0.1, 0.3, 1.0%)<br>CR (0.1, 0.3, 1.0%)<br>SUBJ (0.1, 0.3, 1.0%)<br>MPQA (0.1, 0.3, 1.0%)<br>RT20 (0.1, 0.3, 1.0%) | -0.2, -1.5, -2.1<br>+0.8, 2.4, -0.7<br>-0.2, -1.4, +2.4<br>-0.1, -0.5, -3.3<br>-0.5, +0.4, -0.1<br>+0.2, 2.9, 0.0<br>+2.3, 0.6, -1.9 |
| Intra-LADA [142] | Interpolation of an instance with a randomly reordered version of itself | BERT-base-multilingual-cased + linear layer | CoNLL (5%)<br>CoNLL (100%)<br>GermEval (5%)<br>GermEval (100%) | +0.24 (F1)<br>+0.03 (*)<br>+0.29<br>+0.04 (*) |
| Inter-LADA [142] | Interpolation of the nearest neighbors and sometimes randomly selected instances | BERT-base-multilingual-cased + linear layer | CoNLL (5%)<br>CoNLL (100%)<br>GermEval (5%)<br>GermEval (100%) | +1.32 (F1)<br>+0.64<br>+0.49<br>+0.33 |
| Intra-Inter-LADA [142] | Combination of Intra- and Inter-LADA | BERT-base-multilingual-cased + linear layer | CoNLL (5%)<br>CoNLL (30%)<br>GermEval (5%)<br>GermEval (30%) | +1.57 (F1)<br>+0.59<br>+0.53<br>+0.78 |

* Included in the pretraining

Similarly, Chen et al. [142] propose an interpolation augmentation method in which the hidden layer representations of two samples are interpolated. However, they noticed that this method is not suitable for sequence tagging tasks. For this reason, they propose Intra- and Inter-LADA. Intra-LADA aims to reduce noise from interpolating unrelated sentences by only interpolating an instance with a randomly reordered version of itself. This way, they can increase the performance in every tested task (see Table 8). However, Chen et al. [142] also hypothesize that their Intra-LADA algorithm is limited in producing diverse examples. This limitation leads to Inter-LADA, which sets a trade-off between noise and regularization by interpolating instances that are close together. The closeness is estimated through kNN based on sentence-BERT [143] embeddings and extended by occasional sampling of two completely random instances. As it can be seen in Table 8, Inter-LADA oftentimes performs better than Intra-LADA. The combination of both can further improve the results.

### 3.3 Combination of Augmentation Methods

In augmentation research, a common technique is to combine several data augmentation methods to achieve more diversified instances [144]. Here, combination can mean either the application of multiple separate or stacked methods. For the first kind, Sun and He [51] propose word-level and phrase-level methods that they apply separately. While the results of the word-level and phrase-level methods differ insignificantly, the combination of both groups of methods produced very good results. Similarly, Li et al. [41] combined their proposed methods, which led to better results for the in-domain evaluations. In the work of Bonthu et al. [145] round-trip translation, random swap, random deletion, and random synonym insertion are separately combined, which leads to the best improvement of a LSMT classifier. Furthermore, in contrastive learning, it makes sense to use more than one data augmentation strategy since the goal is to learn meaningful representations that can be fostered by many different views. For example, Yan et al. [146] and Wu et al. [49] use several simple methods of data augmentation for the contrastive learning objective. Details on contrastive learning and the results of the works can be found in the next section. The method of stacking data augmentation techniques, on the other hand, is not always feasible. It is, for example, in most cases not possible to first apply a feature space method and then a data space method. Qu et al. [78] tested this with round-trip translation, cutoff, and adversarial examples. Round-trip translation and the training with adversarial examples produced the best results. Marivate and Sefara [64] stack round-trip translation, synonym and embedding replacement with mixup. In two out of three evaluation settings, this procedure reduces the minimal error.

For the combination of augmentation methods, the meta-learning augmentation approach by Ratner et al. [81] is also of interest. It describes the utilization of a neural network to learn data augmentation transformations [5]. Specifically, Ratner et al. use a GAN to generate sensible sequences of transformations that were defined beforehand. This approach is usable for image as well as text datasets and the authors show that it can achieve a significant improvement when applied to a relation extraction task with augmentations based on language model replacements.

### 3.4 Training Strategies

While semi-supervision is not considered as data augmentation in this work, it can still be sensibly combined through consistency training. In its origin, consistency training is used to make predictions of classifiers invariant to noise [52]. This can be enforced by minimizing the divergences between the output distributions of real and noised instances. Additionally, as only output distributions are included in the process, this consistency can be trained with unlabeled data. Several authors analyze how consistency training behaves when data augmentation methods are used for noise. This process can be illustrated by taking an instance whose label is unknown, applying a label-preserving data augmentation method, and then learning the model to predict the same label for both instances. In this way, the model can learn the invariances and is able to generalize better. Xie et al. [52] show that they achieve very good results by employing consistency training with round-trip translation and a TF-IDF-based replacement method, with an absolute improvement of 22.79% in accuracy on an artificially created low-data regime based on the Amazon-2 dataset with BERT base. They are also able to outperform the state of the art in the IMDb dataset with only 20 supervised instances. Chen et al. [97] even extend this approach within their MixText (TMix) system. First, they generate new instances with round-trip translation. Then, they guess the label of the original and augmented instances by taking a weighted average of the predictions of all of them. In the training, they randomly sample two instances and mix them together with TMix. If one of the two instances is from the original data, they are using the normal supervised loss, but if both instances are from the unlabeled or augmented data, they use the consistency loss, like Xie et al. [52]. Consistency training can also be applied in a supervised fashion as an additional term in the training objective to enforce identical predictions. This is, for example, used in the cutoff method by Shen et al. [101]. They show in their ablation studies that this consistency term improves the accuracy results additively by 0.15%.

Qu et al. [78] combine supervised consistency training with contrastive learning. The contrastive learning scheme should bring the original and augmented instances closer together and the other instances further apart in the representation space. Contrastive learning can be applied in the pretraining phase of a language model so that meaningful representations are learned directly. Wu et al. [49] show that training a language model from scratch with this objective can lead to increased performances for downstream tasks. As augmentation methods the authors use word deletion, span deletion, random reordering and synonym substitution, as well as combinations in sets of two. The evaluation of several tasks shows that there is no clear best augmentation method. Fang et al. [99] and Yan et al. [146] show that contrastive learning can also result in better sentence representations when using an already pretrained model and further training the masked language modeling task with contrastive learning. While the work of Fang et al. [99] uses round-trip translation, Yan et al. [146] experiment with adversarial training, token mixing, cutoff and dropout. Qu et al. [78] and Choi et al. [53] even include contrastive learning into the supervised setting. As augmentation strategies, Qu et al. [78] use adversarial training combined with round-trip translation and Choi et al. [53] use counterfactuals based on language model substitution. Combined with consistency training, Qu et al. [78] achieve even further improvements. A comparing overview can be found in Table 9 of the supplementary material (online).

Other training strategies in which the order of how the training examples are presented to the learning algorithm is altered are for example employed by Liu et al. [129], Yang et al. [147], and Claveau et al. [110]. Liu et al. [129] adopt a curriculum learning approach, where the algorithm first learns less difficult instances. Transferred to the data augmentation topic, the model is first trained with the original data and then with the augmented data. Yang et al. [147] reverse this step and first train the model with the augmented data and then with the original data. This way, the model can correct unfavorable behavior that it learned through noisy augmented data. They also tried an importance-weight loss in which the weights of the synthetic instances are lower but find that the other training method performs better.

### 3.5 Filtering Mechanisms

Mechanisms that filter the generated instances are especially important for methods that are not perfectly label-preserving. A simple mechanism is, for example, employed by Liu et al. [130], who remove generated instances based on the overlap of unigram words with their original equivalents. Similarly, other metrics could also be used, e.g., Levenshtein distance, Jaccard similarity coefficient, or Hamming distance. Wan et al. [131] use a similarity discriminator, initially proposed by Parikh et al. [148], which also measures the similarity of two sentences.

The generative methods by Anaby-Tavor et al. [44], Abonizio and Junior [109], and Claveau et al. [110] filter instances based on a classifier that was trained on the class data. This significantly reduces the diversity of samples, and the classifier cannot really be improved as it is already familiar with these instances. Bayer et al. [46] improve this by using embeddings to measure the quality of the generated instances with regard to the class and more importantly by incorporating the human expert in the loop who needs to determine the correct threshold. However, Yang et al. [147] consider another filtering mechanism in their work which does not require human assistance and is very sophisticated due to the inclusion of two perspectives. Generally, Yang et al. [147] propose a generative method that is suitable for increasing the dataset size for question answering tasks. While they propose to utilize language models for fine-tuning and generation of questions and answers, their filtering methods can be adapted for other data augmentation methods as well. A first filtering mechanism determines whether a generated instance is detrimental by measuring whether the validation loss increases when including the artificial instance. As this would require retraining the model with each generated example, the authors propose to use influence functions [149], [150] to approximate the validation loss change. Furthermore, they first train on the augmented instances and then on the original training data so that the model can adjust itself when unfavorable noise is included in the augmented instances. The other filtering mechanism tries to favor diversity by selecting examples that maximize the number of unique unigrams.

## 4 DISCUSSION: A RESEARCH AGENDA FOR TEXTUAL DATA AUGMENTATION

In the previous section, different data augmentation methods were grouped, explained, compared in terms of performance and put into context with each other. One has to keep in mind that the results reported by the authors of the approaches linked in this survey paper are restricted in their expressiveness and only show one perspective. Many results are limited to special kinds of models and datasets. Based on our findings, we identified an agenda for future research on data augmentation as follows:

### 4.1 Researching the Merits of Data Augmentation in the Light of Large Pre-trained Language Models

Generally, it is not possible to determine which augmentation method works best for a given dataset, nor predict which research direction will be the most appealing in the future. Nevertheless, some patterns in current approaches hint to the directions research can follow in order to overcome current obstacles and challenges. One of the most significant challenges, as formulated by Longpre et al. [4], concerns the usage of large pre-trained language models, which makes the utilization of several data augmentation methods obsolete. Large pre-trained models are currently state of the art, nevertheless, we advise taking further advancements and findings in the research landscape into account, as for example deep belief networks [151], capsule networks [152], or task-specialized networks e.g. for sentiment analysis [153], [154]. Experiments with BERT or other bigger language models are therefore of particular interest. Similarly, several studies [44], [46], [67], [111], [130], [114] have shown that methods only slightly transforming instances with random behavior, such as with synonym replacement (Section 3.1.2.2), EDA (synonym replacement, random swap, deletion, and insertion in one) (Section 3.1.2.1), or by inserting spelling errors (Section 3.1.1.1), tend to be less beneficial in this setting than more elaborate ones. Particularly adversarial training (Section 3.2.1), cutoff (Section 3.2.1), interpolation (Section 3.1.3.2 and 3.2.2), and some generative methods (Section 3.1.4.2) have shown significant improvements with large pre-trained language models. While replacement methods based on embeddings (Section 3.1.2.3) and especially language models (c-BERT) (Section 3.1.2.4) can also gain improvements in combination with those pre-trained models, several studies [44], [64], [78], [111] have shown that the previously mentioned methods can, in most cases, achieve improved results.

The described performance differences become apparent when approaching the challenge highlighted by Longpre et al. [4] from an intuitive perspective. Large language models map data to a latent space with representations nearly invariant to some transformations. For example, synonym replacement methods only replace words that are by definition very close to the representation space, leading to instances that are almost identical [61]. As Longpre et al. [4] hypothesize, data augmentation methods can only be helpful, if they are able to introduce new linguistic patterns. In such instances, using the mentioned methods and generative methods, in particular, might be sensible, as they are based on other large language models that can introduce a high novelty. However, the challenge proposed by Longpre et al. [4] does not have to be universally true. For example, the $SUB^2$ method by Shi et al. [91] only interpolates phrases from the training data and thus does not include unseen linguistic patterns but achieves high gains with a pre-trained model. Another interesting aspect concerns the experiments conducted by Yoo et al. [114], with which they demonstrate that their GPT-3-based generative augmentation method actually improves as the size of the pre-trained classifier increases. The authors hypothesize that larger classifiers have more capacity to better incorporate the GPT3Mix samples.

### 4.2 Improving Existing Data Augmentation Approaches

In general, most promising data augmentation methods have limits and challenges that may be overcome with further research. **Generative models** or their output needs to be conditional on the specific class. Otherwise, the created instances might not preserve the label. This conditioning is oftentimes reached by training a model, which in turn requires enough data to be consistent. Bayer et al. [46] have shown that the conditional model can best replicate the data class, if the problem definition and task data is relatively narrow. Tasks with a broad variety of topics in the data seem less suitable. This problem might be mitigated by adopting new conditioning methods. Currently, most approaches are extended by filter mechanisms. Existing mechanisms, as detailed in Section 3.5, have some drawbacks which might be reduced in the future. Another obstacle concerns generative models themselves, which can require many resources and time to create new instances [46]. Therefore, lightweight alternatives need to be tested in this setting, thus potentially preventing a high dependency on resources, which is referred to by Bayer et al. [46] as the high resource wall problem. Similarly, methods like **round-trip translation** are limited by the underlying model. For example, Marivate and Sefara [64] hypothesize that round-trip translation might not be appropriate for social media data, where translation errors increase. This problem will be addressed in the future, as machine translation models improve their translation capabilities for such difficult instances.

For **adversarial examples**, Liu et al. [129], hypothesize that good generalizability performance stems from the perturbation of the embedding space, rather than the input space. However, data space adversarial training methods should not be disregarded too quickly, as Ebrahimi et al. [21] show that their data space method achieves better results than the virtual adversarial training by Miyato et al. [40]. A general challenge for adversarial training is that it can disturb the true label space in the training data. For example, adversarial example generators often rely on the belief that close input data points tend to have the same labels [121]. Concerning the data space methods, this is often not true for natural language tasks, where few words or even characters determine the class affiliation (e.g., sentiment classification: “I can't believe I like the movie” ⟶small_transformation⟶ “I can' believe I like the movie”). Whether this applies to the adversarial example generators in the feature space needs to be evaluated. If so, research needs to find a way to exclude cases where small transformations disturb labels and at best include cases where stronger transformations still preserve the labels. For this purpose, inspecting feature space methods would be helpful. However, such an inspection is difficult to conduct due to their high-dimensional numerical representation. The same applies to the feature space's **interpolation methods**, where a back transformation to the data space is not trivial. Though, certain approaches, such as those from Liu et al. and Wan et al. [130], [131], use techniques such as encoder-decoder architectures capable of transforming the newly created instances to the data space. An inspection of interpolated instances could lead to interesting insights. This opens another research direction where the **interpolation of instances in the data space** could be further investigated. A method that initially implements this behavior is $SUB^2$ (Section 3.1.3.2), which interpolates instances of the data space through sub-phrase substitutions. This, however, does not result in a high diversity, which is particularly interesting. In this regard, further analysis of the GPT-3 language model by Brown et al. [115] could be valuable, as it shows very interesting interpolation capabilities in the data space.

However, even **avoidably inferior methods** can achieve better results if they are integrated sensibly. The work of Jungiewicz and Pohl [66] can serve as an example. They perform synonym substitution only if it increases the loss of the model. This demonstrates that some data augmentation techniques proposed in the different groups are advanced, sometimes adopting existing methods and refining them.

We highlight some advanced works of the different groups in Table 9 to show which research directions can be considered in the future. It must be emphasized that these methods are not necessarily the best in their groups. The selection is made by the author team on the basis of the information gathered while writing this survey.

Table 9: Collection of some of the most advanced data augmentation techniques for text classification.

| Space | Group | Work | Method description | Improvement |
|---|---|---|---|---|
| Data Space | Character Level Noise | [21] | Flip a letter if it maximizes the loss | +0.62 Acc. (LSTM) |
| | Synonym Replacement | [66] | Only replace words with a synonym if it maximizes the loss | +1.2 Acc. (Kim CNN) |
| | Embedding Replacement | [22] | Choosing embeddings based on the counter-fitting method | -0.6 – +1.9 Acc. (CNN) |
| | | [41] | Counter-fitting, language model selection, and maximizing the prediction probability | Safer model (LSTM) |
| | Language Model Replacement | [67] | c-BERT integrated in reinforcement learning scheme | +0.73 – +1.97 Acc. (BERT) |
| | | [73] | c-BERT and embedding substitution for compound words | +1.9 – +21.0 Acc. (TinyBERT) |
| | Phrase Level Interpolation | [91] | Substitutes substructures | +20.6 – +46.2 Acc. (XLM-R)* |
| | Round-trip Translation | [52] | Random sampling with a temperature parameter | +1.65 Acc. |
| | Generative Methods | [46] | Conditional GPT-2 with human assisted filtering | -2.54 F1 – +15.53 Acc. (ULMFit)* |
| | | [111] | GPT-2 with a reinforcement learning component | +1.0 – +4.3 F1 (XLNet)* |
| Feature Space | Noise | [127] | Virtual adversarial training with special optimization | +0.5 – +5.4 Acc. (RoBERTa-l) |
| | | [129] | Virtual adversarial training with curriculum learning | -0.3 Corr. – +1.2 Acc. (RoBERTa-l) |
| | | [101] | Embedding noising | +0.0 Corr. – +4.4 Acc. (RoBERTa-l) |

| | | | |
|---|---|---|---|
| Interpolation | [138] | Interpolation after last layer of the transformer | -0.01 Acc. – +2.68 Corr. (BERT-l) |
| | [97] | Interpolation of a random BERT layer | +0.0 – +4.6 Acc. (BERT-b)* |
| | [142] | Interpolating neighbors and reordered versions | +0.53 – +1.57 F1 (BERT-b)* |

*Results contain tests on low data regime datasets

### 4.3 Establishing more Comprehensive Evaluation Criteria and Standards for Method Comparison

A general problem in data augmentation research concerns that mostly only improvements with regard to the prediction performance on specific datasets are presented. While this metric is likely the most important one, other metrics, such as time and resource usage, language variety, or configurability, are also important for practitioners as well as for researchers. For example, the generative approaches based on GPT-2 seem very promising when considering prediction performance gain. Nevertheless, language variety is narrowed down, as the model is primarily trained on English data. Furthermore, only few authors discuss the time required for the application of their data augmentation methods. The GPT-2 based method of Bayer et al. [46] takes up to 30 seconds for generating one example, leading to several computing days for a 10-times augmentation of a small dataset. For instance, in the context of crisis informatics this might take too long, as classifiers have to be created quickly for immediate incident management [155]. We therefore urge scientists developing data augmentation techniques to consistently describe the limitations of their approaches. For further data augmentation research, flexible standards should be established in order for methods to be compared more reliably, similar to other machine learning research fields, e.g. few-shot learning [156] or natural language generation [157]. It seems unrealistic that one or few general datasets can capture all peculiarities of data augmentation methods, especially not of those that one tailored to a specific problem. Nevertheless, a small benchmark that can be included in evaluations of upcoming data augmentation methods would be desirable. In the best-case, such a benchmark would address different data augmentation goals, consisting of two or more datasets, from which one replicates a few-shot learning setting and the other a normal learning setting. With the growing usage of generative models, it might also be sensible to consider using datasets that are not part of current training datasets for language models, as an incorporation of testing data would lead to wrong conclusions. The benchmark should not be too large, in order to ensure specific evaluations can still be carried out. Researchers that try to develop such a benchmark, could also consider to specify how much data augmentation should be performed and what models should be used. When determining which model should be used, it might be useful to create an updatable benchmark, as proposed by Gehrmann et al [157], which can be modified according to more recent state-of-the-art models.

### 4.4 Enhancing the Understanding of Text Data Augmentation

Shorten and Khoshgoftaar [5] highlight that while for some image data augmentation techniques it is easy to understand how they might improve the dataset and derived classifiers, however, for other techniques this improvement has not been explainable yet. This also applies to the text regime, where for example, data augmentation methods that paraphrase text without changing the meaning are naturally sensible, while methods applied in the feature space are much more complex to capture. Already the visualization of the data of feature space augmentations created by, for example, adversarial examples or interpolation methods, is much more complicated than in the image domain. As previously elaborated, existing approaches try to convert representations back into the data space by using encoder-decoder architectures [130], [131]. Resulting data space representations could then be investigated and used to better understand underlying data augmentation methods. Furthermore, a more in-depth understanding of why and when data augmentation works needs to be established. With the rise of large language models the question emerges whether data augmentation methods paraphrasing input instances without incorporating new patterns may be obsolete [4]. Certain works have challenged this perspective, by demonstrating that even existing patterns can be beneficial for performance [91]. In this context, it is interesting to note that the augmentation method of Yoo et al. [114] provides better results when the size of the pre-trained language model increases.

### 4.5 Fostering the Usability of Data Augmentation Application

Most data augmentation methods are still research-based in their incremental development progress and therefore not suitable for every practitioner. A simple way to improve the usability is to publish code and, in the best-case, develop libraries that can be used out of the box for augmenting a text dataset. Dhole et al. [158] propose a first large framework to include many text data augmentation methods and filtering mechanisms. The library by Papakipos et al. [159] is not as big for textual data augmentation methods, but can be used for multiple modalities (audio, image, text, & video). While these are very useful libraries, the amalgamation of many procedures comes with abstraction problems. For example, only individual data instances can be transformed and the augmentation procedure does not have access to the entire dataset, so that, for example, no interpolation procedures are implemented. In addition to creating libraries, it might be useful to explore augmentations with a good learning process integration. This can be considered as a criterion to simplify embedding the procedure in the general learning process. Resource utilization, speed, and general continuity in the learning process are crucial for this process. The first two criteria are becoming increasingly relevant as they are related to the current trend of data augmentation, i.e. the use of large underlying models that create a high resource and time execution overhead. As described above, this might be countered with utilizing more lightweight models. A low continuity in the learning process refers to the circumstance that a text data augmentation method is detached from the actual training process; or in the worst case, the learning procedure needs to be split into two halves. The former, also described as offline data augmentation by [37], means that the original data is augmented independently from the model training. A data augmentation technique is called online, if it is embedded into the learning process so that the artificial instances are stochastically included by the learning algorithm, which is, e.g., implemented in the work of Bonthu et al. [145]. The second form occurs, for example, when a feature space method needs to separate the normal network structure, in order to detach the encoder or embedding layer from the rest of the network. This results in a continuity problem of learning, so that, e.g., the encoder or embedding level cannot be trained further.

## 5 CONCLUSION

This survey provides an overview over data augmentation approaches suited for the textual domain. Data augmentation is helpful to reach many goals, including regularization, minimizing label effort, lowering the usage of real-world data particularly in privacy-sensitive domains, balancing unbalanced datasets, and increasing robustness against adversarial attacks (see Section 2). On a high level, data augmentation methods are differentiated into methods applied in the feature and in the data space. These methods are then subdivided into more fine-grained groups, from noise induction to the generation of completely new instances. In addition, we propose several promising research directions that are relevant for future work. Especially in this regard, a holistic view on the current state of the art is necessary. For example, the increasing usage of transfer learning methods makes some data augmentation methods obsolete, as they follow similar goals. Hence, there is a need for more sophisticated approaches that are capable of introducing new linguistic patterns not seen during pre-training, as suggested by Longpre et al. [4].

While data augmentation is increasingly being researched and seems very promising, it also has several limitations. For instance, many data augmentation methods can only create high quality augmented data, if the original amount of data is large enough. Furthermore, as Shorten and Khoshgoftaar [5] describe, data augmentation is not capable of covering all transformation possibilities and eliminating all kinds of biases in the original data. Adopting the example of Shorten and Khoshgoftaar [5], in a news classification task, in which there are no articles containing sports, the standard data augmentation methods will most certainly also not create sport articles, even though this would be necessary. In contrast, data augmentation might induce new undesirable biases. For instance, language models like GPT can contain biases that are then propagated into the dataset [160]. The wide variety of techniques and some very sophisticated methods also bring another layer of complexity that needs to be understood. Moreover, data augmentation can be time consuming, meaning that not all methods are feasible for time critical machine learning development domains, e.g., in some areas of crisis informatics. An increased demand for resources, especially concerning training generative models, is inherent to data augmentation.

In order to mitigate some of the limitations and amplify the strengths of data augmentation, however, we proposed our research agenda, which comprises (1) researching the merits of data augmentation in the light of large pre-trained language models, (2) improving existing data augmentation approaches, (3) establishing more comprehensive evaluation criteria and standards for method comparison, (4) enhancing the understanding of text data augmentation, as well as (5) fostering the usability of data augmentation application.